\documentclass[lettersize,journal]{IEEEtran}
\usepackage{xcolor} 
\usepackage{amsmath, amsfonts, amssymb} 
\usepackage{algorithmic}
\usepackage{algorithm}
\usepackage{array}
\usepackage{textcomp}
\usepackage{stfloats}
\usepackage{url}
\usepackage{verbatim}
\usepackage{graphicx}
\usepackage{cite}
\usepackage{paralist}
\usepackage{adjustbox} 
\usepackage[caption=false,font=normalsize,labelfont=sf,textfont=sf]{subfig}
\usepackage{booktabs}           
\usepackage{multirow}           
\usepackage{rotating}           
\usepackage{siunitx}            
\usepackage{microtype}          
\usepackage[table]{xcolor}      
\usepackage{makecell}
\usepackage{bm}                 
\usepackage{nicefrac}           

\definecolor{linkcolor}{RGB}{255, 0, 150} 
\usepackage[colorlinks=true, allcolors=linkcolor, urlcolor=linkcolor]{hyperref}
\definecolor{UserCyan}{RGB}{0, 191, 255} 

\newcommand{\metricsheader}{%
\multicolumn{1}{c}{$S_m\uparrow$} &
\multicolumn{1}{c}{$F_m\uparrow$} &
\multicolumn{1}{c}{$E_m\uparrow$}}
\newcommand{\best}[1]{\textcolor{red!80!black}{\bfseries #1}}
\newcommand{\secondbest}[1]{\textcolor{UserCyan}{\bfseries #1}}

\newcommand{\fkr}[1]{\textcolor{black}{#1}}

\def\ourmodel{\emph{Samba}}

\begin{document}

\title{Samba+: General and Accurate Salient Object Detection via A More Unified Mamba-based Framework}
\author{
    Wenzhuo Zhao,~
    Keren Fu,~
    Jiahao He,~
    Xiaohong Liu,
    Qijun Zhao,
    and Guangtao Zhai
    
\thanks{Wenzhuo Zhao and Jiahao He are with the College of Computer Science, Sichuan University, Chengdu, 610065, China (Email: zwz@stu.scu.edu.cn, 2023223040091@stu.scu.edu.cn).}
\thanks{Keren Fu and Qijun Zhao are with the College of Computer Science, Sichuan University, China, and are also with the National Key Laboratory of Fundamental Science on Synthetic Vision, Sichuan University, Chengdu, 610065, China (Email: fkrsuper@scu.edu.cn, qjzhao@scu.edu.cn).}
\thanks{Xiaohong Liu is with School of Computer Science, Shanghai Jiao Tong University, Shanghai 200240, China (Email: xiaohongliu@sjtu.edu.cn).}
\thanks{Guangtao Zhai is with the School of Information Science and Electronic Engineering, Shanghai Jiao Tong University, Shanghai 200240, China (email: zhaiguangtao@sjtu.edu.cn).}
}

\maketitle

\begin{abstract}
Existing salient object detection (SOD) models are generally constrained by the limited receptive fields of convolutional neural networks (CNNs) and quadratic computational complexity of Transformers. Recently, the emerging state-space model, namely Mamba, has shown great potential in balancing global receptive fields and computational efficiency. As a solution, we propose Saliency Mamba (\textit{Samba}), a pure Mamba-based architecture that flexibly handles various distinct SOD tasks, including RGB/RGB-D/RGB-T SOD, video SOD (VSOD), RGB-D VSOD, and visible-depth-thermal SOD. Specifically, we rethink the scanning strategy of Mamba for SOD, and introduce a saliency-guided Mamba block (SGMB) that features a spatial neighborhood scanning (SNS) algorithm to preserve the spatial continuity of salient regions. A context-aware upsampling (CAU) method is also proposed to promote hierarchical feature alignment and aggregation by modeling contextual dependencies.
As one step further, to avoid the ``task-specific'' problem as in previous SOD solutions, we develop \textit{Samba+}, which is empowered by training \textit{Samba} in a multi-task joint manner, leading to a more unified and versatile model. Two crucial components that collaboratively tackle challenges encountered in input of arbitrary modalities and continual adaptation are investigated. Specifically, a hub-and-spoke graph attention (HGA) module facilitates adaptive cross-modal interactive fusion, and a modality-anchored continual learning (MACL) strategy alleviates inter-modal conflicts together with catastrophic forgetting. 
Extensive experiments demonstrate that \textit{Samba} individually outperforms existing methods across six SOD tasks on 22 datasets with lower computational cost, whereas \textit{Samba+} achieves even superior results on these tasks and datasets by using a single trained versatile model. Additional results on tasks that emphasize spatial continuity, e.g. camouflaged object detection and skin lesion segmentation further demonstrate the emerging potential of our \textit{Samba} framework for spatially continuous visual modeling.

\end{abstract}


\section{Introduction}
\label{sec:intro}
\IEEEPARstart{S}{alient} object detection (SOD) is an essential vision task that aims to identify and segment the most visually prominent objects within a scene. This technique plays a crucial role in various applications such as object tracking \cite{zhou2021saliency}, semantic segmentation \cite{fu2022light}, image enhancement \cite{miangoleh2023realistic}, autofocus \cite{jiang2024transformer} and evaluation of large models \cite{jiang2024effectiveness}.

Current state-of-the-art (SOTA) SOD methods are primarily dominated by convolutional neural networks (CNNs) and transformers, addressing various SOD tasks, including RGB/RGB-D/RGB-T SOD \cite{wang2023pixels,liu2024vst++,sun2023catnet,hu2024cross,song2022multiple,cong2022does}, video SOD (VSOD) \cite{zhang2021dynamic,zhao2024motion,guo2024unitr}, visible-depth-thermal (VDT) SOD \cite{Song2023HWSI,Wan2024TMNet,Wan2023MFFNet,Luo2024DWFPRNet}, and RGB-D VSOD \cite{mou2023salient,li2024dvsod,lin2024vidsod}. 
While CNN-based backbones are known for their scalability and linear computational complexity, they often suffer from limited receptive fields, making it challenging to capture global dependencies. In contrast, transformer-based backbones offer superior visual modeling by leveraging global receptive fields. However, their self-attention mechanism incurs quadratic complexity, raising efficiency concerns. Although efficient transformer architectures, such as Swin transformer \cite{liu2021swin} and MobileViT \cite{sachin2022mobileviT}, have been proposed, they typically sacrifice some global modeling capability for efficiency, failing to achieve an optimal balance between the two.

\begin{figure}[t!]
    \centering
    \captionsetup{skip=5pt}
    \includegraphics[width=1\linewidth]{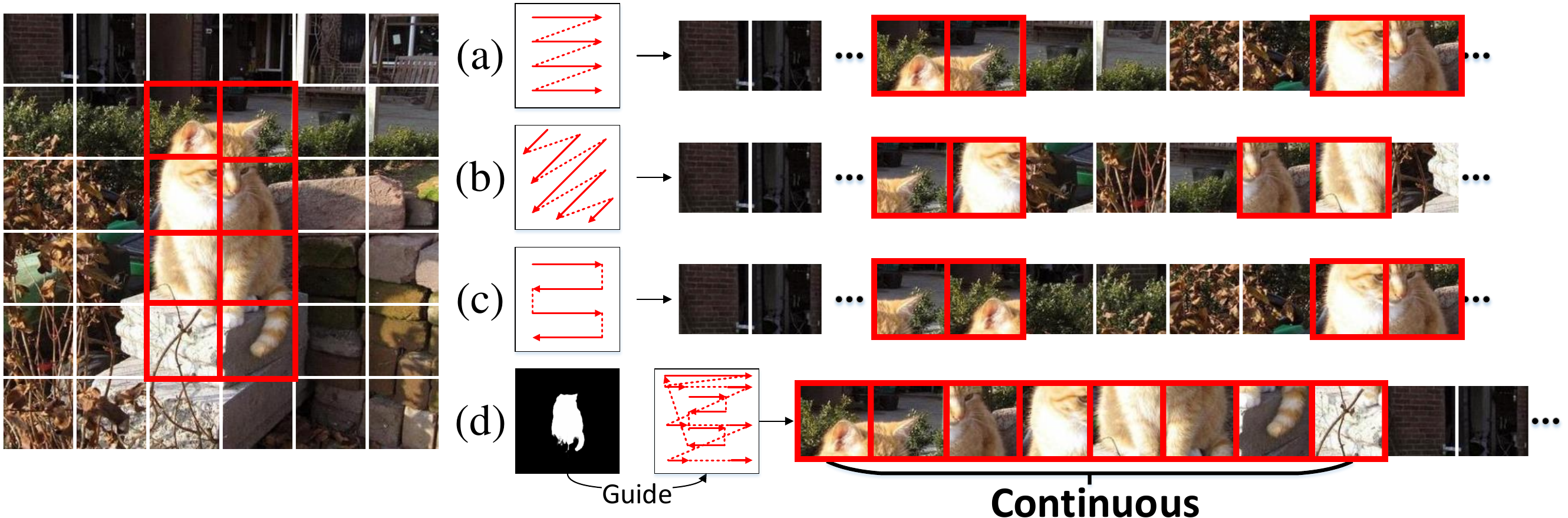}
    \caption{Comparison between existing scanning strategies and our scanning strategy. (a) Sequential scanning of patches in a ``Z'' pattern \cite{liu2024vmamba}. (b) Sequential scanning of patches in diagonal directions \cite{shi2024vmambair,zhao2024rs}. (c) Sequential scanning of patches in an ``S'' pattern \cite{yang2024plainmamba}. (d) Compared to (a/b/c), our spatial neighboring scanning (SNS) can preserve spatial continuity of salient patches.}
    \label{fig_intro}
    \vspace{-0.6cm}
\end{figure}

Recently, Mamba \cite{gu2023mamba}, a novel state space model (SSM), has emerged as a highly promising backbone to balancing global receptive fields and computational efficiency. Mamba employs a selective scanning algorithm to model long-range dependencies while preserving linear complexity. Besides, with a specially designed hardware-aware algorithm, Mamba achieves efficient training on GPUs. Building on this foundation, visual Mamba backbones \cite{zhu2024vim,liu2024vmamba}, with
\fkr{specialized models}\cite{wan2024sigma,yang2024remamber,zhu2024unetmamba,dong2024fusion,zhou2024dmm} based on them, have been rapidly developed. Given Mamba's success across various vision tasks and its absence in SOD areas, we seek its potential for efficient global modeling in SOD tasks.

In this article, we propose two novel unified models, saliency Mamba (\ourmodel) and \textit{Samba+}, to handle general SOD tasks. Owing to the strong performance of visual Mamba backbones, many \fkr{specialized models}\cite{wan2024sigma,zhu2024unetmamba,yang2024remamber,ma2024rs} have leveraged them to extract global cues. Inspired by this, we adopt VMamba \cite{liu2024vmamba} as our backbone, and attempt to design a Mamba-based decoder to produce elaborated results. Nevertheless, we refer to existing Mamba-based decoders \cite{guo2024mambair,zhao2024rs,yang2024plainmamba,wan2024sigma,ma2024rs,xie2024fusionmamba}, and identify two crucial issues of Mamba-based SOD decoders:

\textbf{Spatial continuity issue}. In the workflow of visual Mamba models, 2D feature maps are first divided into patches and then scanned into 1D sequences before being fed to SSMs. 
Since SSMs are designed to provide continuous predictions for 1D causal sequences, the prediction of a current image patch heavily relies on preceding patches, especially the nearest ones. 
Therefore, continuous (or successive) salient patches within 1D sequences can help SSMs accurately locate complete salient regions, enhancing feature representation.
However, previous scanning strategies \cite{liu2024vmamba,shi2024vmambair,zhao2024rs,yang2024plainmamba} neglect this issue and fail to maintain spatial continuity of salient patches (Fig.~\ref{fig_intro} (a), (b) and (c)), hindering SSMs to generate high-quality features. 

\textbf{Feature alignment issue.} Existing Mamba-based decoders \cite{zhu2024unetmamba,ma2024rs,wan2024sigma} typically employ nearest-neighbor interpolation to upsample low-resolution (i.e., high-level) features before incorporating them with high-resolution (i.e., low-level) features. However, this approach leads to two limitations: 1) it lacks learnability; 2) it neglects the contextual dependencies between hierarchical features. These issues result in misalignment during feature fusion, causing deviations in the final prediction. Although previous works \cite{tian2019decoders,liu2023learningto} have proposed learnable upsampling methods, they still fail to model the contextual dependencies between hierarchical features during the upsampling process. 

To address the \textbf{first issue}, we propose a novel saliency-guided Mamba block (SGMB), which emphasizes spatial continuity of salient patches, and leverage SSM's global modeling capability to enhance feature representation. Specifically, we design a spatial neighboring scanning (SNS) algorithm to generate scanning paths, which are applied to flatten 2D feature maps into 1D sequences while preserving spatial continuity of salient patches (Fig.~\ref{fig_intro} (d)). These 1D sequences are then processed by SSMs to generate high-quality features. Notably, compared to commonly fixed scanning strategies (Fig.~\ref{fig_intro} (a/b/c), SNS can dynamically tune scanning directions to handle various scenarios, offering insights for future designs of Mamba scanning strategies. 
To tackle the \textbf{second issue}, we propose a context-aware upsampling (CAU) method, with a novel patch pairing and ordering scheme, to promote hierarchical feature alignment and aggregations during decoding. First, patches from shallow and deep features are paired as subsequences to model the contextual dependencies between hierarchical features. These paired subsequences are then concatenated and input to SSMs. By leveraging powerful causal prediction of SSMs, deep features can progressively learn data distributions of shallow features, and then are expanded to the same shapes as shallow features for fusion. 


\begin{figure}[t!]
    \centering
    \captionsetup{skip=5pt}
    \includegraphics[width=1\linewidth]{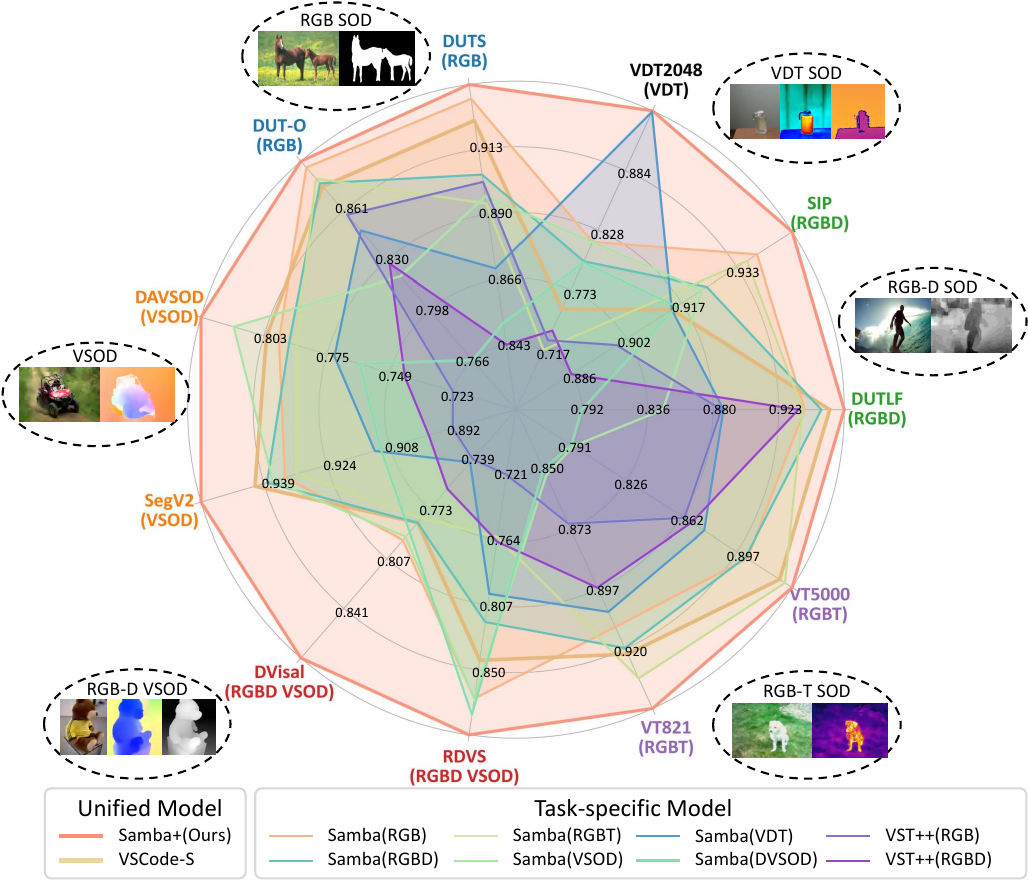}
    \caption{Performance comparisons between our Samba+ and recent state-of-the-art methods (e.g., VSCode\cite{luo2024vscode} and VST++\cite{liu2024vst++}) on various multi-modal SOD tasks, where $S_m$ is the evaluation metric.}
    \label{fig_intro_new}
    \vspace{-0.6cm}
\end{figure}

The above ingredients collectively contribute to \textit{Samba} (our preliminary framework published in CVPR 2025 \cite{He2025Samba}).  
In this article, we have significantly extended the generalizability of \textit{Samba}, overcoming the ``task-specific'' constraint in current SOD techniques. To step toward a truly unified SOD framework, we exploit \textit{Samba+} , a single model jointly trained across diverse SOD tasks plus modality combinations.
To move beyond any task-specific designs and achieve stable joint training across heterogeneous tasks, \textit{Samba+} employs two key components to come across crucial challenges of multi-task \& multi-modality training:
(i) Adaptation to arbitrary input modalities. A hub-and-spoke graph attention (HGA) module is proposed, in which a learnable hub node dynamically aggregates information from multiple-modal spokes, enabling flexible and efficient cross-modal interaction. Instead of relying on task-specific fusion, HGA maintains a unified architecture for both uni-modal and multi-modal tasks, harvesting both parameter efficiency and semantic alignment.
(ii) Stable adaptation for joint learning. We utilize a modality-anchored continual learning (MACL) strategy that enables balanced multi-modal optimization within a Siamese architecture. MACL alleviates inter-modality interference and catastrophic forgetting, thereby ensuring stable learning under both task-incremental and modality-incremental training scenarios.

Fig.~\ref{fig_intro_new} provides a systematic evaluation of \textit{Samba+} against recent state-of-the-art models among various SOD tasks. While task-specific models closely approach the performance ceiling of their respective tasks, generalization to new tasks remains quite limited. In contrast, prompt-based learning and modality adaption \cite{Huang2024Salient,luo2024vscode} represent a promising direction toward universal SOD. However, their performance drops markedly under novel modality combinations (e.g., VDT or RGB-D VSOD), as their reliance on task-fixed prompts restricts their ability to learn modality-invariant representations that capture shared cross-modal saliency. As indicated in Fig.~\ref{fig_intro_new}, \textit{Samba+} generalizes across SOD tasks of heterogeneous modalities via just a single trained versatile model.

In summary, our contributions are as follows:
\begin{compactitem}
    \item
To the best of our knowledge, we are the first to adapt state space models to SOD tasks, and propose a novel unified framework (\textit{Samba}) based on the pure Mamba architecture to flexibly handle general SOD tasks.
    \item
We propose a saliency-guided Mamba block (SGMB), incorporating a spatial neighboring scanning (SNS) algorithm, to maintain spatial continuity of salient patches, thus enhancing feature representation. We also propose a context-aware upsampling (CAU) method to promote hierarchical feature alignment and aggregations by modeling contextual dependencies.
    \item
Building upon \textit{Samba}, we further exploit a more unified model jointly trained across heterogeneous SOD tasks, dubbed \textit{Samba+}, which is also the first truly versatile SOD model in the community. It integrates two essential components, i.e., hub-and-spoke graph attention (HGA) and modality-anchored continual learning (MACL). These two components collaboratively tackle the challenges of modality fusion and stable multi-source joint training.
    \item
\textit{Samba} individually achieves SOTA results across six SOD tasks on 22 datasets, while \textit{Samba+} further generalizes to surpass these results within a single trained model. Its performance on tasks that emphasize spatial continuity (e.g., camouflaged object detection and skin lesion segmentation) highlights the potential of state space modeling as a emerging paradigm for spatially continuous visual modeling. Our results and code are all available at \url{https://github.com/Jia-hao999/Samba}.
\end{compactitem}

The remainder of this article is organized as follows. Section~\ref{sec:relatedwork} discusses related work on various SOD tasks as well as recent advances in visual Mamba models. Section~\ref{sec:method} describes the proposed \textit{Samba} in detail. Section~\ref{sec:Methodology_Samba+} introduces the enhanced version of \textit{Samba}, namely \textit{Samba+}. Experimental results, performance evaluations and comparisons are included in Section~\ref{sec:Experiments}. Finally, conclusions are drawn in Section~\ref{sec:Conclusion}.

\section{Related Work}
\label{sec:relatedwork}
\subsection{Deep Learning based SOD}
\noindent
\textbf{RGB SOD.}
Initially, SOD researches focused solely on the RGB modality, which are dominated by CNN-based architectures. Although CNNs exhibit strong feature extraction capabilities, their limited receptive fields restrict global dependency modeling \cite{wang2021salient}. Subsequent works introduced improvements such as boundary enhancement \cite{zhao2019egnet,ke2022recursive}, feature refinement \cite{zhao2020suppress,wu2022edn} and attention mechanism \cite{zhao2020suppress,wang2023pixels,su2025rapid}. For instance, several methods \cite{zhao2019egnet,wei2020label,ji2020accurate,qin2019basnet,zhang2020select} incorporated explicit edge supervision to promote boundary-aware prediction. Luo et al. \cite{luo2020cascade} aggregated multi-scale context through feature pyramids to better capture global semantics. Attention-based methods \cite{fan2020bbs,zhang2018progressive,liu2018picanet} further improved feature representation by suppressing background noise and enhancing foreground consistency. Recently, transformer-based methods \cite{zhuge2022salient,ma2023boosting,liu2024vst++} have become mainstream for their powerful global modeling and superior performance, yet the quadratic computational complexity of self-attention remains a critical bottleneck. However, these methods face more challenging scenes, such as complex and low-contrast background.

\noindent
\textbf{RGB-D and RGB-T SOD.}
To address the ambiguity and uncertainty that RGB modality often exhibits in challenging scenes, some works introduced depth \cite{fu2020jl,zhou2021specificity,lee2022spsn,sun2023catnet,hu2024cross,huang2021employing,chen20223} or thermal images \cite{tu2021multi,huo2021efficient,cong2022does,chen2022cgmdrnet,zhang2023saliency,zhang2020revisiting} to assist saliency detection. A key challenge in dual-modal SOD lies in effective feature fusion. Early methods \cite{chen2018progressively,chen2019three} typically employed simple fusion strategies such as feature concatenation or element-wise addition. Subsequent works introduced more sophisticated architectures to explicitly model both the consistency and complementarity between modalities. For example, Fu et al. \cite{fu2020jl,fu2022siamese} utilized a Siamese network to extract shared information from RGB and depth inputs for more accurate detection in complex scenes. Tu et al. \cite{tu2021multi} proposed a novel dual-decoder architecture to integrate the multi-type interactions between RGB and thermal. Liu et al. \cite{liu2024vst++} introduced a depth positional encoding that transforms depth information into 3D spatial cues and embeds them into the decoder’s sampling process. \fkr{Tang et al.~\cite{Tangdivide} proposed a divide-and-conquer RGB-T SOD framework that alleviates inter-modality disparities through coordinated modality-specific and complementary feature learning.} Zhang et al. \cite{zhang2025dimsod} further reformulated SOD as a diffusion-based generative task, treating depth and thermal maps as local control signals to unify RGB-D and RGB-T detection within a single framework.

\noindent
\textbf{VSOD.} 
In contrast to static images, video SOD (VSOD) introduces the temporal dimension, posing the significant challenge of jointly modeling motion and appearance cues. To effectively exploit temporal information, existing VSOD methods can be broadly categorized into frame-interaction-based and optical-flow-based approaches. The former \cite{zhao2024motion,wang2017video,fan2019shifting} directly model spatio-temporal dependencies across consecutive frames. For example, Gu et al. \cite{gu2020pyramid} proposed a pyramid aggregation module to integrate multi-scale motion features. Zhang et al. \cite{zhang2021dynamic} dynamically adjusted convolutional filters by learning spatio-temporal representations across frames to adapt to varying scenes and object motions. Recently, Transformer-based models \cite{guo2024unitr} leverage self-attention mechanisms to capture long-range dependencies and exploit cross-modal complementarities. The latter group of methods \cite{liu2023learning,piao2022semi,li2019motion,ji2021full} rely on precomputed optical flow to provide complementary motion cues. Liu et al. \cite{Liu2024Complementary} introduced a flow-guided window attention module to precisely capture long-term spatio-temporal contexts, while Cho et al. \cite{cho2025transflow} leveraged video diffusion models to synthesize realistic and semantically-aware optical flows from static images, enabling motion knowledge transfer for VSOD.

\noindent
\textbf{RGB-D VSOD.}
The effectiveness of integrating depth into VSOD models has been demonstrated in \cite{lu2022depth}, giving rise to a potential research direction: RGB-D VSOD. As acquiring RGB-D videos becomes easier, RGB-D video datasets have been introduced \cite{li2024dvsod,lin2024vidsod,mou2023salient}. At the same time, Mou et al. \cite{mou2023salient} proposed DCTNet+, which leverages a three-stream network focusing on the RGB modality while treating depth and optical flow as auxiliary modalities to effectively utilize the tri-modal information in RGBD-VSOD. Lin et al. \cite{lin2024vidsod} designed a multi-modal integration branch that combines multi-level modal features via a feature aggregation module.

\noindent
\textbf{VDT SOD.}
To overcome the limitations of depth and thermal images, which perform poorly in challenging scenarios and may even degrade the performance of dual-modal SOD methods, Song et al. \cite{Song2023HWSI} constructed the VDT-2048 dataset and proposed a tri-modal (visible-depth-thermal) image fusion detection method. To achieve effective fusion of modal information, some studies \cite{Wan2024TMNet,Wan2023MFFNet} have designed tri-modal interactive encoders to explore the complementarity among the three modalities at the encoding stage; Luo et al. \cite{Luo2024DWFPRNet} designed a modal-selective fusion module that integrates modal features while simultaneously suppressing interference.

\subsection{Visual Mamba}
Motivated by the success of Mamba in language modeling, Zhu et al. \cite{zhu2024vim} transfer this success to vision, and design an efficient visual backbone, vision Mamba (Vim), which incorporates the bidirectional SSMs for global context modeling. Liu et al. \cite{liu2024vmamba} design visual state space blocks, and based on them, a novel visual Mamba backbone (VMamba) is developed and demonstrates promising performance across a range of vision tasks, including semantic segmentation \cite{wan2024sigma,zhu2024unetmamba} and object detection \cite{dong2024fusion,zhou2024dmm}. Due to the impact of the SSM's selective scanning direction on effective receptive fields, Zhao et al. \cite{zhao2024rs} expanded VMamba's four-directional scanning by adding four additional diagonal directions, i.e., Fig.~\ref{fig_intro} (b), to extract large spatial features from multiple directions. Yang et al. \cite{yang2024plainmamba} proposed PlainMamba, with a continuous 2D scanning approach, i.e., Fig.~\ref{fig_intro} (c), ensuring that the scanning sequences are spatially continuous. The potential of the Mamba architecture has recently been demonstrated across multiple vision domains, sparking widespread research interest. For instance, Li et al. \cite{li2025mair} applied it to image restoration, designing a Nested S-shaped Scanning (NSS) strategy to maintain locality and continuity while leveraging sparse sequences to reduce computational costs. In high-level video understanding, Chen et al. \cite{chen2025h} pioneered the use of the Mamba paradigm for hierarchical video understanding in driving scenarios, achieving superior performance on captioning and detection tasks with only one-fifth the computational cost (FLOPs) of Transformers. In video generation tasks, Lu et al. \cite{lu2025end} combined Mamba with multi-step selective diffusion models to balance generation quality and efficiency. However, despite the rapid adoption of Mamba in the aforementioned vision fields, its application in SOD remains largely under-explored.

\section{Methodology of Samba}
\label{sec:method}

\begin{figure*}[!t]
    \centering
    \captionsetup{skip=5pt}
    \includegraphics[width=1\linewidth]{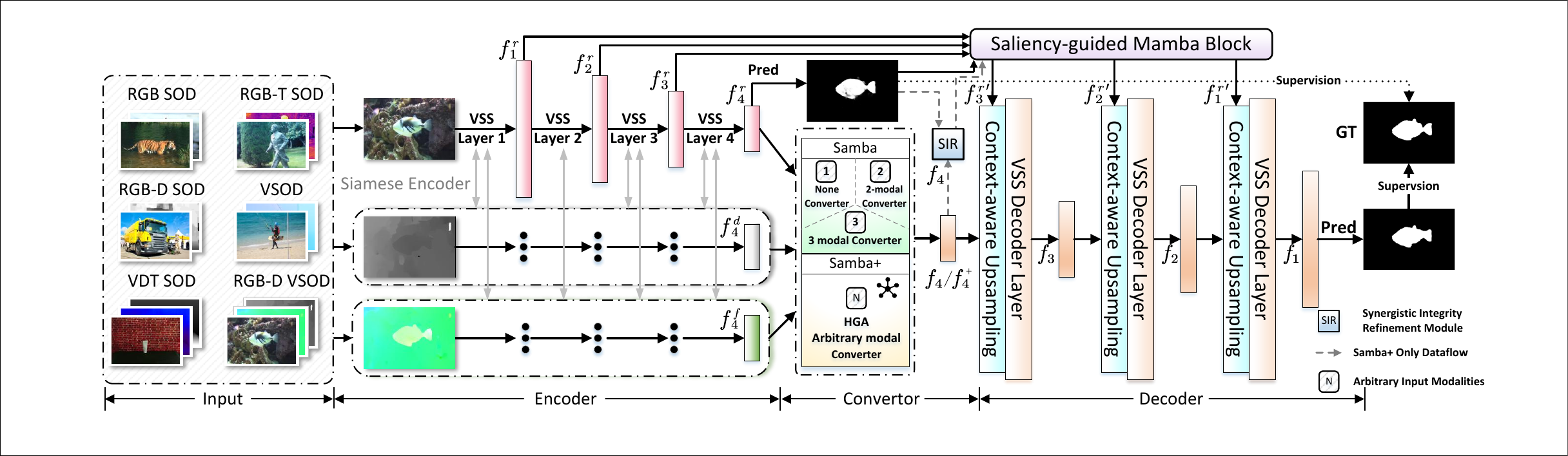}
    \caption{Overall architecture of the proposed \textit{Samba} and  \textit{Samba+} frameworks for general SOD tasks. Modules specific to \textit{Samba+} are indicated in the figure.
    }
    \label{fig_overview}
    \vspace{-4mm}
\end{figure*}

\subsection{Preliminaries}
\label{sec:preliminaries}
SSMs \cite{gu2021combining,gu2021efficiently,smith2022simplified} are sequence-to-sequence models designed to capture long-range dependencies using linear time-invariant (LTI) systems. These systems map an input sequence $x\left( t \right) \in \mathbb{R}$ to an output sequence $y\left( t \right) \in \mathbb{R}$ via a latent state $h\left( t \right) \in \mathbb{R}^\mathcal{N}$, represented by the  ordinary differential equations (ODEs):
\begin{equation}
\label{eq:ssm}
    y(t)={\rm \textbf{C}}h(t)+{\rm \textbf{D}}x(t),~
    h'(t)={\rm \textbf{A}}h(t)+{\rm \textbf{B}}x(t),
\end{equation}
where ${\rm \textbf{A}} \in \mathbb{R}^{\mathcal{N}\times \mathcal{N}}$,~ ${\rm \textbf{B}} \in \mathbb{R}^{\mathcal{N} \times 1}$,~${\rm \textbf{C}} \in \mathbb{R}^{1\times \mathcal{N}}$,~and ${\rm \textbf{D}} \in \mathbb{R}^1$ are system parameters. $h'(t)$ is the time derivative of $h(t)$. \fkr{Conceptually, the state matrix \textbf{A} controls the memory dynamics, whereas \textbf{B} and \textbf{C} determine how information enters and leaves the latent space.} The discretization of this continuous-time system is essential for integrating SSMs into deep learning frameworks. A common method for discretization is zero-order hold (ZOH) \cite{gu2021efficiently}, which can be formulated as:
\begin{equation}
    \overline{\rm \textbf{A}}= {\rm exp}({\rm {\Delta \textbf{A}}}),~
    \overline{\rm \textbf{B}}=({\rm {\Delta \textbf{A}}})^{-1}({\rm exp(\textbf{A})}-\textbf{I})\cdot {\rm \Delta \textbf{B}},
\end{equation}
where $\Delta \in \mathbb{R}^\mathcal{D}$ is a predefined timescale parameter. \fkr{The ZOH scheme discretizes the system by integrating the continuous dynamics over each interval, preserving the temporal behavior encoded in the original ODE.} Such a process allows the discrete-time SSM to be expressed in a recurrent form, mapping the input sequence $\left\{ x_1,x_2,...,x_k \right\}$ to the output sequence $\left\{ y_1,y_2,...,y_k \right\}$:
\begin{equation}
\label{eq:discret-ssm}
    y_k={\rm \textbf{C}}h_k+{\rm \textbf{D}}x_k,~
    h_k=\overline{\rm \textbf{A}}h_{k-1}+\overline{\rm \textbf{B}}x_k.
\end{equation}

SSMs efficiently handle long-range dependencies with linear complexity, but their time-invariance limits capturing dynamic context. To address this, the recent advancement, Mamba \cite{gu2023mamba}, introduce an input-dependent selection mechanism (S6) that relaxes the time-invariance constraint. Mamba allows the system parameters, specifically $\textbf{B}$, $\textbf{C}$, $\Delta$, to vary based on the input, thereby enhancing the model's flexibility in capturing complex interactions within long sequences. Conceptually, the selective mechanism in Mamba modulates the effective SSM parameters in a data-dependent manner, allowing the model to adjust its state transitions based on the input. This input-driven modulation enables the model to better capture non-stationary or context-dependent temporal patterns beyond what fixed LTI dynamics can represent. Mamba also employs a new parallel scanning algorithm, enabling efficient training on GPUs.

\subsection{Framework Overview}
In this section, we present an overview of the proposed \ourmodel~model for general SOD tasks, as shown in Fig.~\ref{fig_overview}. The content of \textit{Samba+} will be introduced in next Sec.~\ref{sec:Methodology_Samba+}. The ``Input'' encompasses various SOD tasks. The ``Encoder'' employs a Siamese backbone that contains four visual state space (VSS) layers \cite{liu2024vmamba}, to extract multi-level features from the inputs. The ``converter'' integrates information from different modalities through a multi-modal fusion Mamba (MFM). For the ``Decoder'', we mainly propose a novel saliency-guided Mamba block (SGMB) and a context-aware upsampling (CAU) method. SGMB employs a novel scanning strategy to maintain spatial continuity of salient patches, thereby enhancing feature representation. CAU is designed to facilitate the alignment and aggregations of hierarchical features by modeling contextual dependencies. In the subsequent sections, we provide detailed descriptions of the ``Encoder'', ``Converter'', and ``Decoder''.

\subsection{Encoder}
\label{sec:encoder}
\begin{figure}[t!]
    \centering
    \captionsetup{skip=5pt}
    \includegraphics[width=1\linewidth]{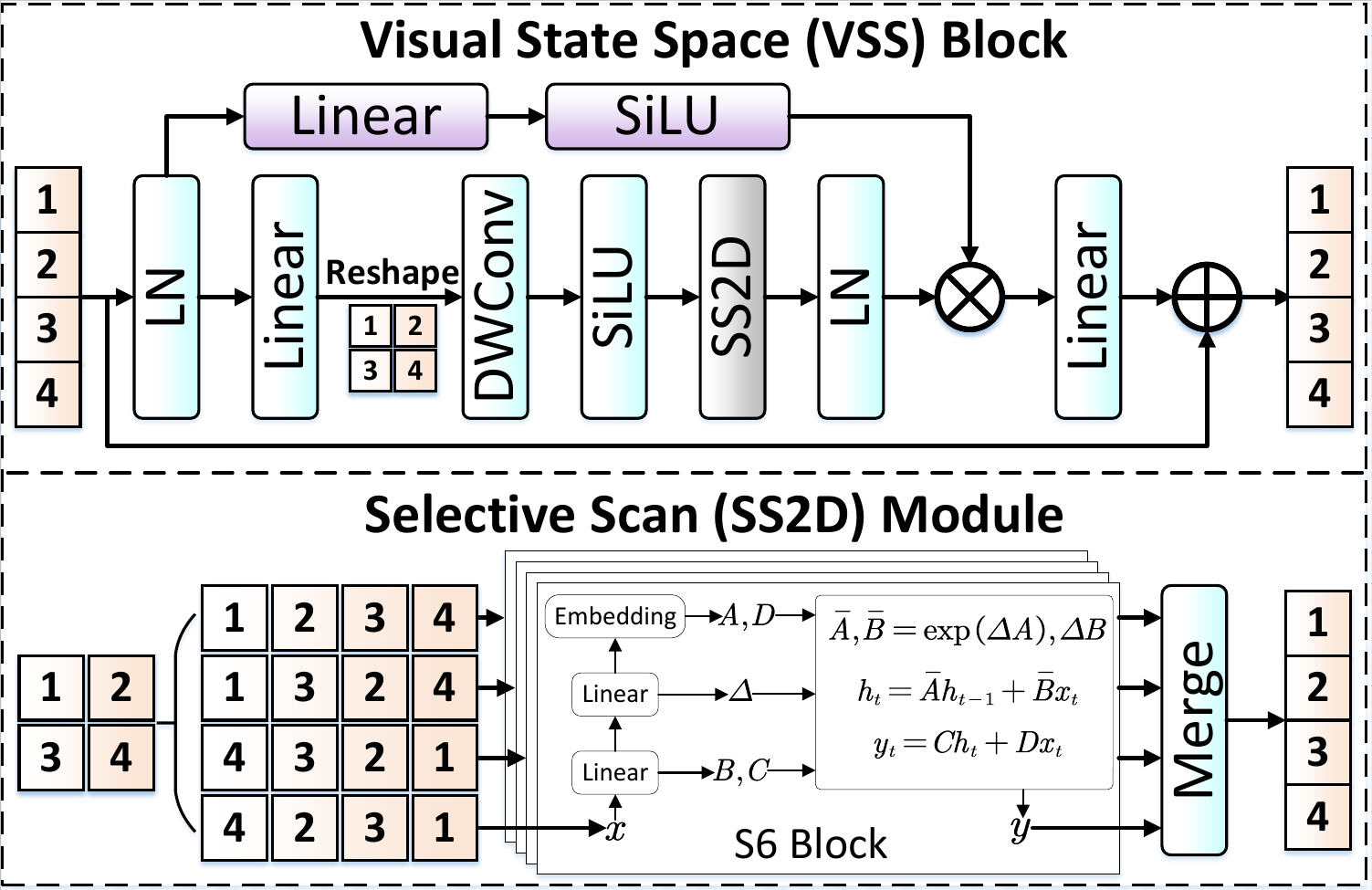}
    \caption{Diagram of the VSS block and SS2D module.}
    \label{fig_vss}
    \vspace{-0.2cm}
\end{figure}

To address general SOD tasks, we implement a Siamese encoder based on VSS layers. The encoder begins by partitioning the input images into patches. Then four VSS layers, each containing multiple VSS blocks and a downsampling operation, are cascaded to extract multi-level features $f_{i}^{m}$, where $m\in \left[ r,d,f,t \right]$ represent RGB, depth, optical flow and thermal modalities, respectively, and $i\in \left[ 1,2,3,4 \right]$ denotes the layer index. Fig.~\ref{fig_vss} provides a detailed illustration of the VSS block and its core selective scan (SS2D) module.

\noindent
\textbf{VSS.} The input first undergoes a layer normalization (LN), after which it is split into two information flows. The first flow is processed by a sequence of operations: a linear projection (Linear), a reshape operation (Reshape), a depth-wise convolution (DWConv) and a SiLU activation function \cite{elfwing2018sigmoid}. Next, an SS2D module is applied to model global dependencies, followed by another LN layer. The second flow, by contrast, only passes through a Linear and a SiLU activation function. After that, the two flows are multiplied and processed by a Linear layer for an output. Finally, the original input is added to the output via a residual connection.

\noindent
\textbf{SS2D.} The input 2D feature is first flattened into four 1D sequences by scanning along four distinct directions. Then the four sequences are processed by S6 blocks \cite{gu2023mamba} to capture long-range dependencies. 
Lastly, the sequences are reordered into the same direction, and then summed to merge information.

\vspace{-1mm}
\subsection{Converter}
\vspace{-1mm}
\label{sec:converter} 
To facilitate flexible extension from single-modal SOD (RGB SOD) to dual-modal (RGB-D/T SOD, VSOD) and tri-modal SOD (RGB-D VSOD, VDT SOD), we introduce a multi-modal fusion Mamba (MFM) module as a converter between the encoder and decoder. When handling the RGB SOD task, the converter remains empty, and $f_{4}^{r}$ is directly fed to the decoder. Within the dual-modal converter, $f_{4}^{r}\in \mathbb{R} ^{H_{4}\times W_{4}\times C_{4}}$ and $f_{4}^{x}\in \mathbb{R} ^{H_{4}\times W_{4}\times C_{4}}$, where $x\in \left[ d,f,t \right]$, are first processed by a Linear and a DWConv, respectively. The outputs are then flattened into $\mathbb{R} ^{L\times C_{4}}$, where $L=H_{4}\times W_{4}$, and concatenated along the $L$ dimension. To explore the interaction of multi-modal information, we utilize an S6 block to process the concatenated sequence. Finally, the sequence is split into two outputs, which are summed and processed by a Linear projection. This process can be formulated as:
\begin{equation}
\begin{aligned}
    &\bar{f}_{4}^{i} = DWConv\!\left( Linear\!\left( f_{4}^{i} \right) \right), \quad i \in \{r,x\},\\
    &\tilde{f}_{4}^{r},\tilde{f}_{4}^{x}
      = Split\!\left( S6\!\left( Cat\!\left( \bar{f}_{4}^{r},\bar{f}_{4}^{x} \right) \right) \right),\\
    &f_{4}=Linear\!\left( \tilde{f}_{4}^{r}+\tilde{f}_{4}^{x} \right).
\end{aligned}
\end{equation}
Building upon the dual-modal converter, we can seamlessly extend it to a tri-modal converter by continuing to increase the number of modal processing branches.

\begin{algorithm}[t]
\caption{Core process of SNS}
\small
\begin{algorithmic}[1]
\label{alg:sns}
\renewcommand{\algorithmicrequire}{\textbf{Input:}}
\renewcommand{\algorithmicensure}{\textbf{Output:}}
\REQUIRE{2D coarse saliency map ${S}_{c}$ with shape = $(h, w)$}
\ENSURE{1D array $I_s$ that stores the indexes of all salient patches}
\STATE Initialize the current scanning row of ${S}_{c}$ as the first row ($cur=1$), the scanning direction of $cur$ from left to right ($dir=l\rightarrow r$) and 1D array $I_s$ as empty ($I_s=\varnothing$).
\WHILE{$cur<=h$}
    \STATE Scan the salient patches in the current row according to the direction $dir$, and append their indexes to $I_s$.
    \STATE Calculate the distances between the last salient patch in $I_s$ and both the leftmost (${dist}_{left}$) and rightmost (${dist}_{right}$) salient patches in the next row.
    \IF{${dist}_{left}<={dist}_{right}$}
        \STATE Set the scanning direction from left to right ($l\rightarrow r$).
    \ELSE
    \STATE Set the scanning direction from right to left ($r\rightarrow l$).
    \ENDIF
    \STATE Jump to the next row($cur=cur+1$).
\ENDWHILE
\STATE Return $I_s$.
\end{algorithmic}
\end{algorithm}

\begin{figure}[t!]
    \centering
    \captionsetup{skip=5pt}
    \includegraphics[width=1\linewidth]{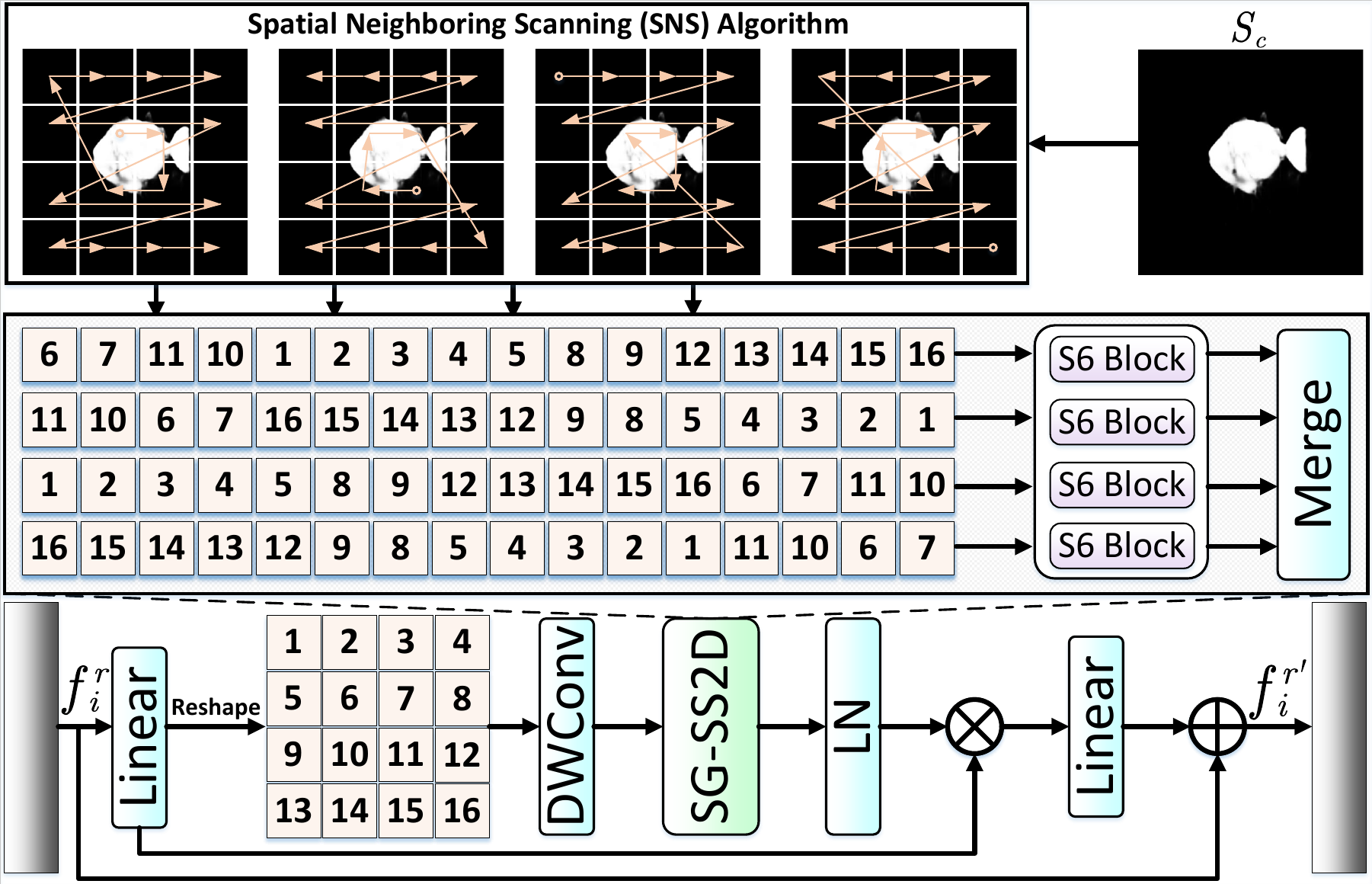}
    \caption{Diagram of the saliency guided Mamba block (SGMB).}
    \label{fig_sgm}
    \vspace{-0.3cm}
\end{figure}

\begin{figure}[t!]
    \centering
    \captionsetup{skip=5pt}
    \includegraphics[width=1\linewidth]{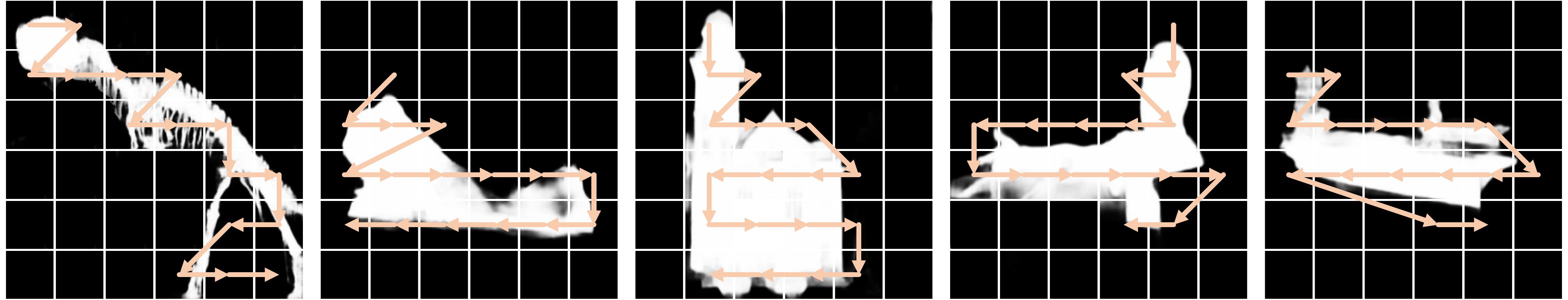}
    \caption{Scanning paths of salient regions generated by SNS.}
    \label{fig_sns}
    \vspace{-0.6cm}
\end{figure}

\subsection{Decoder}
\label{sec:decoder}
As discussed in Sec.~\ref{sec:intro}, two crucial issues remain in designing Mamba-based SOD decoders. To address them, we propose a novel saliency-guided Mamba block (SGMB) and a context-aware upsampling (CAU) method.

\subsubsection{Saliency-guided Mamba Block}
\label{sec:saliency-guided}
As shown in Fig.~\ref{fig_overview}, the extracted RGB features $f_{i}^{r}$, where $i\in \left[ 1,2,3 \right]$, and a coarse saliency map $S_{c}$ predicted from $f_{4}^{r}$ are fed to SGMB, aiming to enhance the RGB features. As emphasized in Sec.~\ref{sec:intro}, maintaining spatial continuity of salient patches within 1D sequences is crucial for accurate prediction by SSM. To this end, we design a spatial neighboring scanning (SNS) algorithm that flattens 2D feature maps into 1D sequences while preserving spatial continuity of salient patches, as illustrated in Fig.~\ref{fig_sgm}. 
Specifically, we transfer maintaining spatial continuity of salient patches to a shortest path traversal problem of salient patches. In other words, during scanning, the next salient patch to be scanned should be spatially close to the current salient patch. 
In order to reduce computational complexity and keep the traversal path as short as possible, SNS scans each salient patch through an approximate shortest path. 
The core process of SNS is detailed in Algorithm~\ref{alg:sns}.

The input is $S_c$, where each salient patch is assigned an index representing its position in the map. The output is a list $I_s$ that stores the indexes of all salient patches, reflecting the scanning path of salient regions. Starting from the first row, SNS scans all salient patches row by row. Since salient patches within the same row are almost continuous, we can approximate that two adjacent patches within the same row are the closest to each other. Thus, we can scan the salient patches within each row either from left to right or right to left. After scanning the current row, the algorithm moves to the next row. To maintain spatial continuity and minimize computation time, we compare the distance between the last salient patch in the current row and both the leftmost and rightmost salient patches in the next row. The patch with the smaller distance is selected as the starting patch in the next row. These steps are repeated until all rows have been scanned, and the final $I_s$ is generated. 
Once $I_s$ is obtained, we store the indexes of all non-salient patches sequentially in a list $I_{ns}$, and then concatenate it to $I_s$, generating a complete scanning path for 2D feature maps. To enhance the robustness of SNS, we generate three additional variants of the scanning path by altering the directions: (1) concatenate $I_s$ to $I_{ns}$; (2) reverse $I_s$, $I_{ns}$ and concatenate $I_{ns}$ to $I_s$; (3) reverse $I_s$, $I_{ns}$ and concatenate $I_s$ to $I_{ns}$. These scanning paths are then applied to the RGB features, flattening them into 1D sequences. The remaining steps of SGMB are similar to those in the VSS block. Finally, the input RGB features $f_{i}^{r}$ are enhanced into high-quality features ${f_{i}^{r}}^{\prime}$.

\begin{figure}[t!]
    \centering
    \captionsetup{skip=5pt}
    \includegraphics[width=1\linewidth]{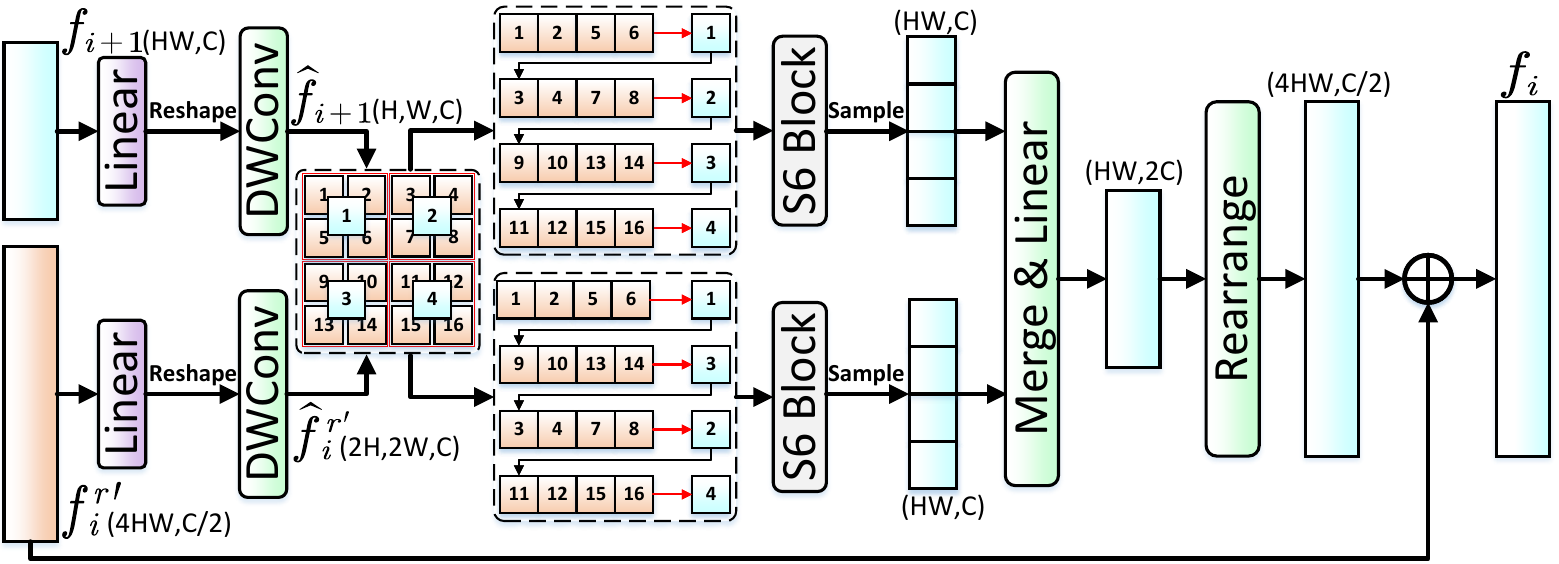}
    \caption{Diagram of the context-aware upsampling (CAU) method.}
    \label{fig_cau}
    \vspace{-0.6cm}
\end{figure}

\noindent
\textbf{Discussions of SNS.} \textbf{(1)} Compared to the ``S'' pattern in Fig. ~\ref{fig_intro}, our SNS requires extra computation to determine the scanning direction for each row to preserve spatial continuity. However, in certain cases, SNS could generate the similar scanning paths of salient regions as the ``S''-like pattern. Thus, to confirm the necessity of SNS, we count the number of images where scanning paths of salient regions are different from the ``S'' pattern, and its proportion across all images is found to be $\sim$31\%. Some visual examples are provided in Fig.~\ref{fig_sns}. \textbf{(2)} Notably, from another perspective, the proposed SNS can be deemed as a novel approach to inject information of a prior map into the processed features, by changing patch orders according to the map. Such an approach could provide insights for future designs of Mamba-based infrastructures.


\subsubsection{Context-aware Upsampling Method}
\label{sec:context-aware}

Previous upsampling methods lack learnability and fail to model the contextual dependencies between hierarchical features, leading to misalignment during feature fusion. To tackle this, we propose a learnable context-aware upsampling (CAU) method, and integrate it into the decoder. 

Fig.~\ref{fig_cau} shows the diagram of CAU. Firstly, two input features $f_{i+1}\in \mathbb{R} ^{HW\times C}$ and ${f_{i}^{r}}^{\prime}\in \mathbb{R} ^{4HW\times {C/2}}$, where $i\in \left[ 1,2,3 \right]$, are processed by a Linear, a Reshape and a DWConv, yielding $\hat{f}_{i+1}\in \mathbb{R} ^{H\times W\times C}$ and $\hat{f}_{i}^{r\prime}\in \mathbb{R} ^{2H\times 2W\times C}$. To align $\hat{f}_{i+1}$ with $\hat{f}_{i}^{r\prime}$ after upsampling, we propose leveraging the patch information of $\hat{f}_{i}^{r\prime}$ to guide the upsampling process. Due to the neighborhood correlation of patches between hierarchical features, each patch in $\hat{f}_{i+1}$ can be associated with four most relevant patches within $\hat{f}_{i}^{r\prime}$. Based on this, we can model the contextual dependencies between $\hat{f}_{i+1}$ and $\hat{f}_{i}^{r\prime}$. Specifically, we first group the patches of $\hat{f}_{i}^{r\prime}$ using $2\times 2$ windows. Then, we sample the patches from $\hat{f}_{i+1}$ one by one, and sequentially pair them with the patch groups of $\hat{f}_{i}^{r\prime}$. Next, we concatenate the paired subsequences into a long sequence, and input it to an S6 block. By exploiting the causal prediction capabilities of the S6 block, each patch from $\hat{f}_{i+1}$ can progressively simulates the feature distribution of its corresponding patch group from $\hat{f}_{i}^{r\prime}$. To enhance this simulation process, we alter the connection order of the paired subsequences to generate a new long sequence, and input it to the other S6 block. Afterward, we sample the patches belonging to $\hat{f}_{i+1}$ from the processed long sequences, and restore them to the same order, yielding two features shaped as $\mathbb{R} ^{HW\times C}$. Then they are merged (summed) and processed by a Linear layer to generate new features shaped as $\mathbb{R} ^{HW\times 2C}$. To reorganize its feature distribution, we use a rearrange function to expand its length fourfold and reduce its channels to a quarter of the original, obtaining upsampled features shaped as $\mathbb{R} ^{4HW\times {C/2}}$. Finally, the original ${f_{i}^{r}}^{\prime}$ is added to the upsampled features for feature aggregation.

\subsubsection{VSS Decoder Layers}
\label{sec:vss-decoder}
We implement VSS decoder layers based on VSS blocks, aiming to decode the aggregated features from CAU. To explore inter-channel dependencies, we introduce a channel attention mechanism (CAM) \cite{hu2018squeeze} following SS2D, forming our VSS decoder layers. The layer process mainly follows a LN $\rightarrow $ Linear $\rightarrow$ DWConv $\rightarrow$ SS2D $\rightarrow$ CAM $\rightarrow$ LN $\rightarrow$ Linear flow with a residual connection.

\subsection{Loss Function for Samba}
We adopt a combination of widely used binary cross entropy (BCE) loss and intersection-over-union (IoU) loss for training our \ourmodel, which is formulated as:
\begin{equation}
\label{for5}
    \mathcal{L} =\mathcal{L} _{bce}+\mathcal{L} _{iou}. 
\end{equation}
Our total loss is defined as: 
\begin{equation}
    \mathcal{L} _{total}=\mathcal{L} \left( S_c,GT \right) +\mathcal{L} \left( S_f,GT \right),
\end{equation} 
where $GT$ represents ground truth, $S_c$ represents the coarse saliency map predicted by $f_{4}^{r}$, and $S_f$ represents the final output saliency map of \ourmodel.

\begin{figure*}[!t]
    \centering
    \captionsetup{skip=5pt}
    \includegraphics[width=1\linewidth]{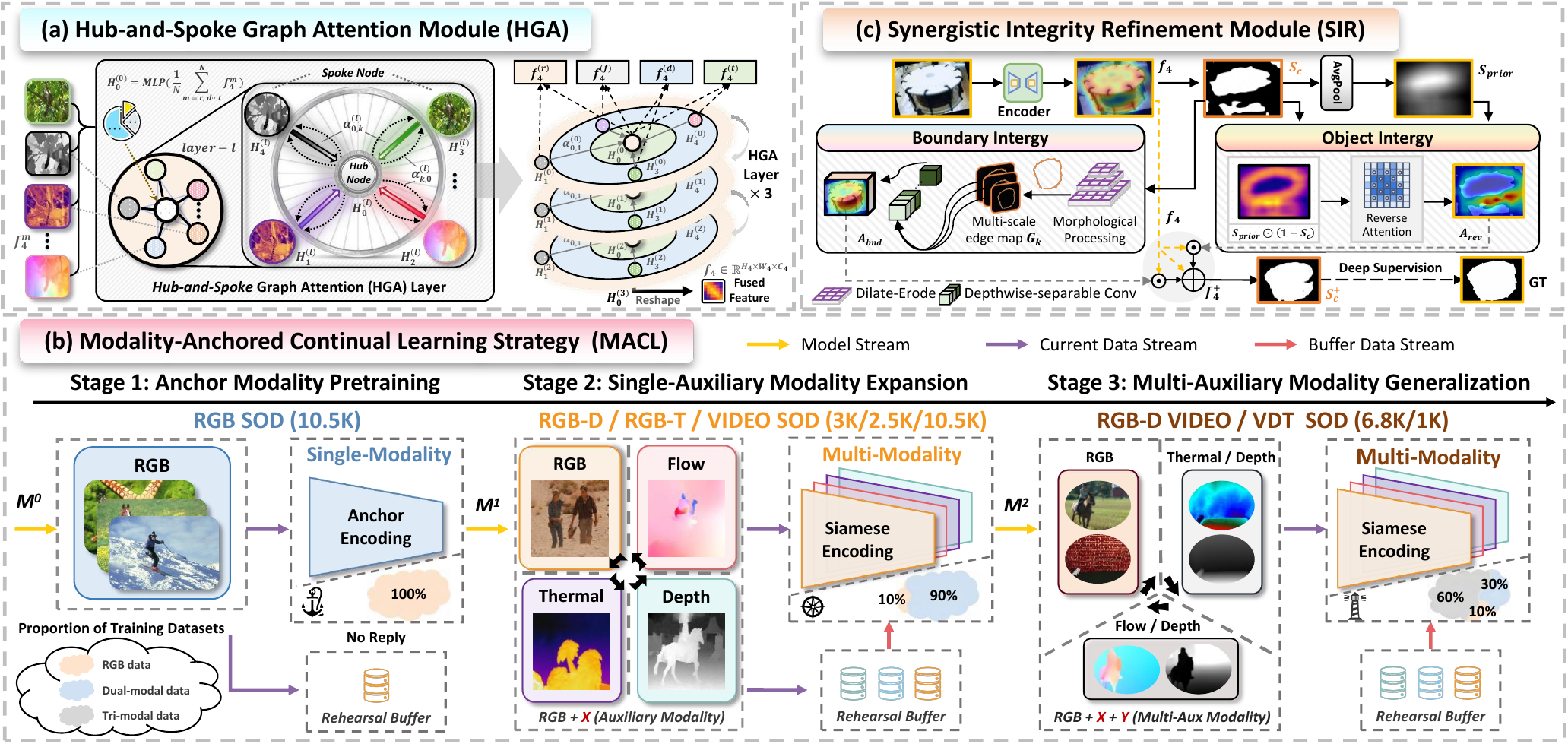}
    \caption{Three crucial components that collaboratively resolve the challenges faced by \textit{Samba+} model for general SOD tasks.}
    \label{fig_overview_new}
    \vspace{-6mm}
\end{figure*}

\section{Methodology of Samba+}
\label{sec:Methodology_Samba+}

\subsection{Framework Overview}
\textit{Samba+} is built upon the \textit{Samba} framework (Fig.~\ref{fig_overview}), which integrates SGMB, the SNS mechanism, and the CAU-based decoder introduced in~Sec. \ref{sec:method}. Furthermore, \textit{Samba+} incorporates two major and one minor modifications. The former two include a hub-and-spoke graph attention (HGA) converter and a modality-anchored continual learning (MACL) strategy. The latter one is a synergistic integrity refinement (SIR) module. Note that HGA and SIR are also reflected in Fig.~\ref{fig_overview}, while MACL is only given in Fig.~\ref{fig_overview_new} since it is the training strategy for the architectural framework.  

For the rationale of HGA (Fig.~\ref{fig_overview} and Fig.~\ref{fig_overview_new}~(a)), the aforementioned \textit{Samba} converter (i.e. MFM) is tailored to specific SOD tasks (namely that it in practice leads to distinct parametric designs for different SOD tasks), requiring separate training for each modality configuration and hindering effective unification across heterogeneous inputs. Meanwhile, recent studies on multi-modal Mamba models \cite{li2025alignmamba} indicate that heterogeneous modality statistics disrupt the reliability of Mamba’s sequential state-space scanning and limit the capture of cross-modal dependencies, thereby necessitating a modality-agnostic converter capable of harmonizing heterogeneous encoder features.
The above concerns lead to our HGA converter that aligns arbitrary combinations of modalities into a unified latent space. 

Regarding MACL (Fig.~\ref{fig_overview_new}~(b)), if naive multi-task joint training is applied, \textit{Samba+} would process heterogeneous modalities alternately, causing competition in its shared recurrent state. As discussed in \cite{liu2025vision}, Mamba’s hidden state is sensitive to frequent shifts across tasks. Such instability drives modality-specific gradients to interfere, weakening the model's ability to preserve task-relevant information and ultimately exacerbating cross-task interference \cite{long2024dgmamba}. This interference resembles the optimization in continual learning, where sequential tasks disrupt prior representations. As a result, inspired by versatile self-supervised learning~\cite{ye2024continual}, we reconsider the joint training of heterogeneous SOD tasks from an incremental-alignment perspective, and propose MACL to alleviate inter-modal conflicts and catastrophic forgetting.

Further for SIR (Fig.~\ref{fig_overview} and Fig.~\ref{fig_overview_new}~(c)), as joint training with heterogeneous data source introduces distributional discrepancies and cross-domain noise, weakening the spatial continuity of coarse map \cite{Vu_2019_CVPR}, we design SIR to optimize spatial integrity of the coarse map, facilitating accurate SNS scanning.

\subsection{Hub-and-Spoke Graph Attention (HGA)}

In HGA (Fig.~\ref{fig_overview_new}~(a)), a learnable hub node serves as the central pivot, while each modality is represented as a spoke node. When only one modality is given, HGA naturally degenerates into a self-attention operation, ensuring consistent architecture and efficient parameter usage in both uni-modal and multi-modal settings. Specifically, we first aggregate the multi-modality features $\{f_{4}^{r},f_{4}^{x}\}$ using an MLP (Linear $\rightarrow$ ReLU $\rightarrow$ Linear) to obtain the permuted initial hub feature map
$\mathbf{H}_{\text{0}}^{0} \in \mathbb{R}^{C_{4} \times H_{4} \times W_{4}}$, where the subscript ``$0$'' means the hub node. All modality feature maps are redefined as $\mathbf{H}^{0}_{k} \in \mathbb{R}^{C_{4} \times H_{4} \times W_{4}}, k \in \{1,\ldots,N\}$, where $N$ is the total modality number. Based on this initialization, $L$ hub-and-spoke graph layers built upon GATv2~\cite{brody2021graph} are utilized to process interactions across the $N$ modalities. For the hub map $\mathbf{H}_{0}$ and spoke maps $\mathbf{H}_{k}$, the update rule at layer $l \in \{1,2,..,L\}$ ($L$ means the total layer iterations) is defined:
\vspace{-3pt}
\begin{equation}
\mathbf{H}_{0}^{l} = \sigma \left(
\alpha_{0,0}^{(l-1)} \odot \mathbf{W}^{l}\mathbf{H}_{0}^{(l-1)} 
+ \sum_{k=1}^{N} \alpha_{0,k}^{(l-1)} \odot \mathbf{W}^{l}\mathbf{H}_{k}^{(l-1)}
\right),
\label{eq:hub_update}
\end{equation}
where $\mathbf{W}^{l}$ is a learnable $1\times1$ convolution shared across all nodes, and $\odot$ denotes element-wise multiplication with broadcast along channels. $\sigma$ is the ReLU function, and ${\alpha}_{0,0}^{(l-1)}$, ${\alpha}_{0,k}^{(l-1)} \in \mathbb{R}^{1 \times H_{4} \times W_{4}}$ denote the attention logits of the hub node and spoke nodes:
\vspace{-3pt}
\begin{equation}
\alpha_{0,k}^{(l-1)} = 
\frac{\exp\!\left(
\mathrm{\sigma}\!\left(
\mathbf{a}^{\top}
\big[
\mathbf{W}^{l}\mathbf{H}_{0}^{(l-1)}, \mathbf{W}^{l}\mathbf{H}_{k}^{(l-1)}
\big]
\right)
\right)}
{\sum_{j=0}^{N}
\exp\!\left(
\mathrm{\sigma}\!\left(
\mathbf{a}^{\top}
\big[
\mathbf{W}^{l}\mathbf{H}_{0}^{(l-1)}, \mathbf{W}^{l}\mathbf{H}_{j}^{(l-1)}
\big]
\right)
\right)},
\label{eq:attention_coeff_merged}
\end{equation}
where $[\cdot , \cdot] \in \mathbb{R}^{2C_4 \times H_{4} \times W_{4}}$ denotes channel-wise concatenation, $\mathbf{a}\in\mathbb{R}^{2C_4}$ is a shared attention vector implemented as a $1\times1$ convolution mapping $2C_4 \rightarrow 1$. This GATv2 formulation~Eq. \ref{eq:attention_coeff_merged} outputs pixel-wise attention coefficients $\alpha_{0,k}^{(l-1)}$ that act as adaptive gating factors. Unlike conventional fully connected graphs, spoke nodes only update their representations by absorbing feedback from the hub with logits ${\alpha}_{k,0}^{(l-1)}$ via the same mechanism, fostering deep bidirectional modal interaction. After $L$ HGA layers ($L=3$ is set in practice), the final hub $\mathbf{H}_{0}^{L}$ is taken as the converter output, providing the aggregated multi-modal representation $f_4$. 


\subsection{Modality-Anchored Continual Learning (MACL)}


To alleviate inter-modal conflicts together with catastrophic forgetting, MACL (Fig.~\ref{fig_overview_new} (b)) adopts a progressive three-stage schedule from the incremental-alignment perspective:

\noindent\textbf{Stage-1: Anchor modality pretraining.} \textit{Samba+} adopts a VMamba encoder pretrained on RGB images. We align it to SOD by training on RGB datasets, establishing an RGB latent representation that serves as the ``anchor'' model. By deferring multi-modal interaction in early training, we avoid gradient conflicts in low-level layers and obtain stable feature space. 

\noindent\textbf{Stage-2: Single-auxiliary modality expansion.}
Built upon the anchored RGB representation, we handle ``RGB + X (one randomly auxiliary modality)'' via Siamese encoder.  As directly training with auxiliary modalities may cause the shared encoder to drift away from the RGB anchor established in Stage-1, we construct a rehearsal buffer to replay data from the previous stage.  10\% of the training samples are drawn from the rehearsal buffer for experience replay. This acts as an anchor-preserving constraint, allowing auxiliary modalities to contribute complementary cues without disrupting the stable latent structure of the RGB representation.

\noindent\textbf{Stage-3: Multi-auxiliary modality generalization.} In the final stage, more complex tri-modal configurations are introduced. At the same time, dual-modal datasets from the previous stage are added to the rehearsal buffer for experience replay. Specifically, RGB samples are replayed at a rate of 10\% to preserve the anchor space, while dual-modality samples are oversampled at 30\% to address the imbalance in rehearsal from earlier stages. This systematic experience replay enables the Siamese encoder to integrate all modalities effectively without compromising the RGB anchor, resulting in a consistent multi-modal representation across heterogeneous SOD tasks.

Note that during training, multi-modal data imbalance caused by disparate dataset sizes and limited non-RGB augmentation results in biased optimization of \textit{Samba}' Siamese encoder. To mitigate this imbalance, we apply the Randomized Quantization (RQ)~\cite{Wu2023Randomized} to augment non-RGB modalities. Since RQ can introduce perturbations that may degrade the physical properties of these modalities, we constrain the perturbation magnitude within salient regions using the corresponding saliency masks.

\subsection{Synergistic Integrity Refinement (SIR)}
Inspired by integrity learning~\cite{zhuge2022salient}, SIR collaboratively optimizes spatial integrity of a coarse map at both object and boundary levels, as illustrated in Fig.~\ref{fig_overview_new} (c). In contrast to \textit{Samba}, SIR operates on the fused bottleneck feature $f_{4} \in\mathbb{R}^{H_{4}\times W_{4}\times C_{4}}$ and its induced coarse map $S_c$, and consists of two complementary parts:

\noindent\textbf{(1) Boundary integrity.}
To enhance boundary consistency, we compute multi-scale soft morphological edge map of $S_c$ under kernel size $k\in\{3,5,7\}$:
\begin{equation}
G_k = \mathrm{Dilate}(S_c,k)-\mathrm{Erode}(S_c,k).
\end{equation}
All edge responses are concatenated along channel dimension and transformed into boundary channel attention tensor $A_{\text{bnd}}\in\mathbb{R}^{H_{4}\times W_{4}\times C_{4}}$ by using a depthwise-separable convolution (DSConv).

\noindent\textbf{(2) Object integrity.}
To restore macroscopic completeness, we first obtain a coarse global shape prior $S_{\text{prior}}$ by applying a $14\times14$ large average pooling to $S_c$. Regions that are globally consistent with the foreground shape but locally underestimated are recovered via the below reverse attention:
\begin{equation}
R = S_{\text{prior}}\odot(1-S_c).
\end{equation}
A DSConv further transforms $R$ into an object-guided channel attention tensor $A_{\text{rev}}
\in\mathbb{R}^{H_{4}\times W_{4}\times C_{4}}$. 
Finally, the boundary and object attention tensors jointly modulate the bottleneck feature: ${f}_{4}^{+}
= f_{4} + f_{4} \odot  A_{\text{bnd}} + f_{4} \odot  A_{\text{rev}}$.
The integrity-enhanced feature ${f}_{4}^{+}$ then generates a spatially refined coarse map ${S}_c^{+}$, which is together fed to the \textit{Samba} decoder instead of $S_c$.

\subsection{Loss Function for \textit{Samba+}}
To tackle the imbalance between foreground and background in multi-task training, we extend the \textit{Samba} loss (Eq.~\ref{for5}) by incorporating a structure-aware spatial weighting factor~$w$ that is derived from GT \cite{He2023FEDER,he2025run}, along with an additional focal loss~\cite{lin2017focal}. 
\fkr{Empirically, such modifications yield $\sim$1\% gains on some hard datasets.} 
Thus, the modified loss is defined as:
\begin{equation}
\mathcal{L}
= \mathcal{L} _{bce}^{w}+\mathcal{L} _{iou}^{w}+\mathcal{L} _{focal}.
\end{equation}

\section{Experiments and Results}
\label{sec:Experiments}
\begin{table*}[t]
  \definecolor{subheadercolor}{gray}{0.92} 
\definecolor{ourscolor_1}{RGB}{255, 245, 235} 
\definecolor{ourscolor_2}{RGB}{254, 230, 210} 
  \centering
  \small 
  \caption{Quantitative comparison of our \textit{Samba} and \textit{Samba+} against other SOTA RGB SOD methods on five benchmark datasets. ``-'' indicates the result is not available. The best and second-best results are highlighted in \textcolor{red}{red} and \secondbest{blue}, respectively.}
  \vspace{-2mm}
  \renewcommand{\arraystretch}{0.9} 
  \setlength\tabcolsep{2.45pt} 
  \begin{tabular}{l|cc|ccc|ccc|ccc|ccc|ccc}
    \toprule
    \multirow{2.5}{*}{Method} & \multirow{2.5}{*}{\makecell{Params \\ (M)}}  & \multirow{2.5}{*}{\makecell{MACs \\ (G)}} & \multicolumn{3}{c}{DUTS\!\cite{wang2017learning}}  & \multicolumn{3}{c}{DUT-O\!\cite{yang2013saliency}} & \multicolumn{3}{c}{HKU-IS\!\cite{li2015visual}} & \multicolumn{3}{c}{PASCAL-S\!\cite{li2014secrets}} & \multicolumn{3}{c}{ECSSD\!\cite{yan2013hierarchical}} \\
    \cmidrule(lr){4-6} \cmidrule(lr){7-9} \cmidrule(lr){10-12} \cmidrule(lr){13-15} \cmidrule(lr){16-18}
    & & & $S_m\uparrow$ & $F_m\uparrow$ & $E_m\uparrow$ & $S_m\uparrow$ & $F_m\uparrow$ & $E_m\uparrow$ & $S_m\uparrow$ & $F_m\uparrow$ & $E_m\uparrow$ & $S_m\uparrow$ & $F_m\uparrow$ & $E_m\uparrow$ & $S_m\uparrow$ & $F_m\uparrow$ & $E_m\uparrow$\\
    \midrule
    \multicolumn{18}{c}{\cellcolor{subheadercolor}\textbf{\textit{CNN-based Methods}}} \\
    \midrule
    \text{GateNet-R}$_{20}$ \cite{zhao2020suppress} & 128.63 & 162.22 & 0.891 & 0.874 & 0.932 & 0.840 & 0.782 & 0.878 & 0.921 & 0.926 & 0.959 & 0.863 & 0.836 & 0.886 & 0.924 & 0.935 & 0.955 \\
    \text{CSF-R2}$_{20}$ \cite{gao2020highly} & 36.53 & 18.96 & 0.890 & 0.869 & 0.929 & 0.838 & 0.775 & 0.869 & - & - & - & 0.863 & 0.839 & 0.885 & 0.931 & 0.942 & 0.960\\
    \text{EDN}$_{22}$ \cite{wu2022edn} & 42.85 & 20.41 & 0.892 & 0.893 & 0.933 & 0.849 & 0.821 & 0.884 & 0.924 & 0.940 & 0.963 & 0.864 & 0.879 & 0.907 & 0.927 & 0.950 & 0.957 \\
    \text{ICON-R}$_{22}$ \cite{zhuge2022salient} & 33.09 & 20.91 & 0.890 & 0.876 & 0.931 & 0.845 & 0.799 & 0.884 & 0.920 & 0.931 & 0.960 & 0.862 & 0.844 & 0.888 & 0.928 & 0.943 & 0.960 \\
    \text{MENet}$_{23}$ \cite{wang2023pixels} & 27.83 & 94.62 & 0.905 & 0.895 & 0.943 & 0.850 & 0.792 & 0.879 & 0.927 & 0.939 & 0.965 & 0.871 & 0.848 & 0.892 & 0.927 & 0.938 & 0.956\\
    \midrule
    \multicolumn{18}{c}{\cellcolor{subheadercolor}\textbf{\textit{Transformer-based Methods}}} \\
    \midrule
    \text{EBM}$_{21}$ \cite{zhang2021learning} & 118.96 & 53.38 & 0.909 & 0.900 & 0.949 & 0.858 & 0.817 & 0.900 & 0.930 & 0.943 & 0.971 & 0.877 & 0.856 & 0.899 & 0.941 & 0.954 & 0.972 \\
    \text{ICON-S}$_{22}$ \cite{zhuge2022salient} & 94.30 & 52.59 & 0.917 & 0.911 & 0.960 & 0.869 & 0.830 & 0.906 & 0.936 & 0.947 & 0.974 & 0.885 & 0.860 & 0.903 & 0.941 & 0.954 & 0.971 \\
    \text{BBRF}$_{23}$ \cite{ma2023boosting}  & 74.40 & 48.60 & 0.908 & 0.905 & 0.951 & 0.855 & 0.820 & 0.898 & 0.935 & 0.946 & 0.936 & 0.871 & 0.884 & 0.925 & 0.939 & 0.957 & 0.972 \\
    \text{VST-S\texttt{++} }$_{24}$ \cite{liu2024vst++} & 74.90 & 32.73 & 0.909 & 0.897 & 0.947 & 0.859 & 0.813 & 0.890 & 0.932 & 0.941 & 0.969 & 0.880 & 0.859 & 0.901 & 0.939 & 0.951 & 0.969 \\
    \text{VSCode-S}$_{24}$ \cite{luo2024vscode} & 74.72 & 93.76 & 0.926 & 0.922 & 0.960 & 0.877 & 0.840 & 0.912 & 0.940 & 0.951 & 0.974 & 0.887 & 0.864 & 0.904 & 0.949 & 0.959 & \secondbest{0.974} \\
    \midrule
    \rowcolor{ourscolor_1}
    \textit{Samba (Ours)} & 49.59 & 46.68 & \secondbest{0.932} & \secondbest{0.930} & \secondbest{0.966} & \secondbest{0.889} & \secondbest{0.859} & \secondbest{0.922} & \secondbest{0.945} & \secondbest{0.956} & \secondbest{0.978} & \secondbest{0.892} & \secondbest{0.896} & \secondbest{0.931} & \secondbest{0.953} & \secondbest{0.965} & \best{0.978} \\
    \rowcolor{ourscolor_2}
    \textit{Samba+ (Ours)} & 59.03 & 48.46 & \best{0.936} & \best{0.933} & \best{0.968} & \best{0.893} & \best{0.860} & \best{0.927} & \best{0.947} & \best{0.957} & \best{0.979} & \best{0.896} & \best{0.898} & \best{0.935} & \best{0.955} & \best{0.966} & \best{0.978} \\
    \bottomrule
  \end{tabular}
  \vspace{-4mm}
  \label{RGB_SOTA}
\end{table*}

\begin{table*}[t]
  \definecolor{subheadercolor}{gray}{0.92} 
  \definecolor{ourscolor}{gray}{0.92}      
  \definecolor{ourscolor_1}{RGB}{235, 250, 240} 
  \definecolor{ourscolor_2}{RGB}{210, 240, 220} 
  \centering
  \small 
  \caption{Quantitative comparison of our \textit{Samba} and \textit{Samba+} against other SOTA RGB-D SOD methods on five benchmark datasets. The best and second-best results are highlighted in \textcolor{red}{red} and \secondbest{blue}, respectively.}
  \vspace{-2mm}
  \setlength\tabcolsep{2.3pt} 
  \renewcommand{\arraystretch}{0.9} 
  \begin{tabular}{l|cc|ccc|ccc|ccc|ccc|ccc}
    \toprule
    \multirow{2.5}{*}{Method} & \multirow{2.5}{*}{\makecell{Params \\ (M)}}  & \multirow{2.5}{*}{\makecell{MACs \\ (G)}} & \multicolumn{3}{c}{STERE\!\cite{niu2012leveraging}} & \multicolumn{3}{c}{DUTLF-D\!\cite{piao2019depth}} & \multicolumn{3}{c}{SIP\!\cite{fan2020rethinking}} & \multicolumn{3}{c}{NLPR\!\cite{peng2014rgbd}} & \multicolumn{3}{c}{NJUD\!\cite{ju2014depth}}  \\
    \cmidrule(lr){4-6} \cmidrule(lr){7-9} \cmidrule(lr){10-12} \cmidrule(lr){13-15} \cmidrule(lr){16-18}
    & & & $S_m\uparrow$ & $F_m\uparrow$ & $E_m\uparrow$ & $S_m\uparrow$ & $F_m\uparrow$ & $E_m\uparrow$ & $S_m\uparrow$ & $F_m\uparrow$ & $E_m\uparrow$ & $S_m\uparrow$ & $F_m\uparrow$ & $E_m\uparrow$ & $S_m\uparrow$ & $F_m\uparrow$ & $E_m\uparrow$\\
    \midrule
    \multicolumn{18}{c}{\cellcolor{subheadercolor}\textbf{\textit{CNN-based Methods}}} \\
    \midrule
    \text{BBSNet}$_{20}$ \cite{fan2020bbs} & 49.77 & 31.20 & 0.908 & 0.903 & 0.942 & 0.882 & 0.870 & 0.912 & 0.879 & 0.884 & 0.922 & 0.931 & 0.918 & 0.961 & 0.921 & 0.919 & 0.949 \\
    \text{JL-DCF}$_{20}$ \cite{fu2020jl} & 143.52 & 211.06 & 0.900 & 0.895 & 0.942 & 0.894 & 0.891 & 0.927 & 0.885 & 0.894 & 0.931 & 0.931 & 0.918 & 0.965 & 0.877 & 0.892 & 0.941 \\
    \text{SP-Net}$_{21}$ \cite{zhou2021specificity} & 67.88 & 175.29 & 0.907 & 0.906 & 0.949 & 0.895 & 0.899 & 0.933 & 0.894 & 0.904 & 0.933 & 0.927 & 0.919 & 0.962 & 0.925 & 0.928 & 0.957 \\
    \text{DCF}$_{21}$ \cite{ji2021calibrated} & 53.92 & 108.60 & 0.906 & 0.904 & 0.948 & 0.925 & 0.930 & 0.956 & 0.874 & 0.886 & 0.922 & 0.922 & 0.910 & 0.957 & 0.904 & 0.905 & 0.943 \\
    \text{SPSN}$_{22}$ \cite{lee2022spsn} & - & - & 0.907 & 0.902 & 0.945 & - & - & - & 0.892 & 0.900 & 0.936 & 0.923 & 0.912 & 0.960 & 0.918 & 0.921 & 0.952 \\
    \midrule
    \multicolumn{18}{c}{\cellcolor{subheadercolor}\textbf{\textit{Transformer-based Methods}}} \\
    \midrule
    \text{SwinNet-B}$_{21}$ \cite{liu2021swinnet} & 199.18 & 122.20 & 0.919 & 0.918 & 0.956 & 0.918 & 0.920 & 0.949 & 0.911 & 0.927 & 0.950 & 0.941 & 0.936 & 0.974 & 0.920 & 0.924 & 0.956 \\
    \text{CATNet}$_{23}$ \cite{sun2023catnet} & 262.73 & 172.06 & 0.920 & 0.922 & 0.958 & 0.952 & 0.958 & 0.975 & 0.910 & 0.928 & 0.951 & 0.938 & 0.934 & 0.971 & 0.932 & 0.937 & 0.960 \\
    \text{VST-S\texttt{++}}$_{24}$ \cite{liu2024vst++} & 143.15 & 45.41 & 0.921 & 0.916 & 0.954 & 0.945 & 0.950 & 0.969 & 0.904 & 0.918 & 0.946 & 0.935 & 0.925 & 0.964 & 0.928 & 0.928 & 0.957 \\
    \text{CPNet}$_{24}$ \cite{hu2024cross} & 216.50 & 129.34 & 0.920 & 0.922 & \secondbest{0.960} & 0.951 & 0.959 & 0.974 & 0.907 & 0.927 & 0.946 & 0.940 & 0.936 & 0.971 & 0.935 & 0.941 & 0.963 \\
    \text{VSCode-S}$_{24}$ \cite{luo2024vscode} & 74.72 & 93.76 & 0.931 & 0.928 & 0.958 & \secondbest{0.960} & \secondbest{0.967} & \secondbest{0.980} & 0.924 & 0.942 & 0.958 & 0.941 & 0.932 & 0.968 & 0.944 & 0.949 & 0.970 \\
    \midrule
    \rowcolor{ourscolor_1}
    \textit{Samba (Ours)} & 54.94 & 71.64 & \secondbest{0.935} & \secondbest{0.933} & \best{0.963} & 0.956 & 0.964 & 0.976 & \secondbest{0.931} & \secondbest{0.949} & \secondbest{0.966} & \secondbest{0.947} & \secondbest{0.941} & \best{0.976} & \secondbest{0.949} & \best{0.956} & \best{0.975} \\
    \rowcolor{ourscolor_2}
    \textit{Samba+ (Ours)} & 59.03 & 72.50 & \best{0.937} & \best{0.935} & \best{0.963} & \best{0.967} & \best{0.974} & \best{0.984} & \best{0.948} & \best{0.960} & \best{0.977} & \best{0.949} & \best{0.944} & \secondbest{0.975} & \best{0.950} & \secondbest{0.954} & \secondbest{0.973} \\
    \bottomrule
  \end{tabular}
  \vspace{-4mm}
  \label{RGBD_SOTA}
\end{table*}

\begin{table*}[t]
  \definecolor{subheadercolor}{gray}{0.92}
  \definecolor{ourscolor_1}{RGB}{255, 250, 235} 
  \definecolor{ourscolor_2}{RGB}{255, 245, 210} 

  \centering
  \small 
  \caption{Quantitative comparison of our \textit{Samba} and \textit{Samba+} against other SOTA VSOD methods on five benchmark datasets. The best and second-best results are highlighted in \textcolor{red}{red} and \secondbest{blue}, respectively.}
  \vspace{-2mm}
  \setlength\tabcolsep{2.2pt} 
  \renewcommand{\arraystretch}{0.9} 
  
  \begin{tabular}{l|cc|ccc|ccc|ccc|ccc|ccc}
  
    \toprule
    \multirow{2.5}{*}{Method} & \multirow{2.5}{*}{\makecell{Params \\ (M)}}  & \multirow{2.5}{*}{\makecell{MACs \\ (G)}} & \multicolumn{3}{c}{DAVSOD-easy\!\cite{fan2019shifting}} & \multicolumn{3}{c}{SegV2\!\cite{li2013video}} & \multicolumn{3}{c}{DAVIS\!\cite{perazzi2016benchmark}} & \multicolumn{3}{c}{FBMS\!\cite{ochs2013segmentation}} & \multicolumn{3}{c}{VOS\!\cite{li2017benchmark}} \\
    
    \cmidrule(lr){4-6} \cmidrule(lr){7-9} \cmidrule(lr){10-12} \cmidrule(lr){13-15} \cmidrule(lr){16-18}
    & & & $S_m\uparrow$ & $F_m\uparrow$ & $E_m\uparrow$ & $S_m\uparrow$ & $F_m\uparrow$ & $E_m\uparrow$ & $S_m\uparrow$ & $F_m\uparrow$ & $E_m\uparrow$ & $S_m\uparrow$ & $F_m\uparrow$ & $E_m\uparrow$ & $S_m\uparrow$ & $F_m\uparrow$ & $E_m\uparrow$\\
    \midrule
    \multicolumn{18}{c}{\cellcolor{subheadercolor}\textbf{\textit{CNN-based Methods}}} \\
    \midrule
    
    \text{MGAN}$_{19}$ \cite{li2019motion} & 91.51 & 123.57 & 0.740 & 0.611 & 0.778 & 0.902 & 0.869 & 0.950 & 0.913 & 0.894 & 0.965 & 0.909 & 0.903 & 0.946 & 0.797 & 0.725 & 0.829 \\
    \text{PCSA}$_{20}$ \cite{gu2020pyramid} & - & - & 0.725 & 0.590 & 0.759 & 0.886 & 0.848 & 0.938 & 0.900 & 0.877 & 0.960 & 0.872 & 0.844 & 0.917 & 0.802 & 0.699 & 0.816 \\
    \text{FSNet}$_{21}$ \cite{ji2021full} & 83.41 & 35.32 & 0.760 & 0.637 & 0.796 & 0.849 & 0.773 & 0.920 & 0.922 & 0.909 & 0.972 & 0.875 & 0.867 & 0.918 & 0.678 & 0.621 & 0.755 \\
    \text{DCFNet}$_{21}$ \cite{zhang2021dynamic} & 69.56 & 93.27 & 0.729 & 0.612 & 0.781 & 0.903 & 0.870 & 0.953 & 0.914 & 0.899 & 0.970 & 0.883 & 0.853 & 0.910 & 0.838 & 0.773 & 0.861 \\
    \text{UGPL}$_{22}$ \cite{piao2022semi} & - & - & 0.732 & 0.602 & 0.771 & 0.867 & 0.828 & 0.938 & 0.911 & 0.895 & 0.968 & 0.897 & 0.884 & 0.939 & 0.751 & 0.685 & 0.811 \\
    \text{MMNet}$_{24}$ \cite{zhao2024motion} & 50.81 & 82.63 & 0.732 & 0.602 & 0.771 & 0.867 & 0.828 & 0.938 & 0.911 & 0.895 & 0.968 & 0.897 & 0.884 & 0.939 & 0.751 & 0.685 & 0.811 \\
    \midrule
    \multicolumn{18}{c}{\cellcolor{subheadercolor}\textbf{\textit{Transformer-based Methods}}} \\
    \midrule
    \text{MGTNet}$_{22}$ \cite{min2022mutual} & 150.91 & 265.21 & 0.765 & 0.653 & 0.800 & 0.903 & 0.861 & 0.946 & 0.925 & 0.919 & 0.976 & 0.900 & 0.881 & 0.929 & 0.814 & 0.727 & 0.819 \\
    \text{CoSTFormer}$_{23}$ \cite{liu2023learning} & - & - & 0.779 & 0.667 & 0.819 & 0.874 & 0.813 & 0.943 & 0.923 & 0.906 & 0.978 & 0.869 & 0.861 & 0.913 & 0.791 & 0.708 & 0.811 \\
    \text{UFO}$_{24}$ \cite{guo2024unitr} & 55.92 & 248.80 & 0.747 & 0.626 & 0.799 & 0.888 & 0.850 & 0.951 & 0.918 & 0.906 & 0.978 & 0.858 & 0.868 & 0.911 & - & - & - \\
    \text{VSCode-S}$_{24}$ \cite{luo2024vscode} & 74.72 & 93.76 & 0.800 & 0.710 & 0.835 & \secondbest{0.946} & 0.937 & 0.984 & 0.936 & 0.922 & 0.973 & 0.905 & 0.902 & 0.939 & - & - & - \\
    \midrule
    \rowcolor{ourscolor_1}
    \textit{Samba (Ours)} & 54.94 & 71.64 & \secondbest{0.813} & \secondbest{0.734} & \secondbest{0.856} & 0.943 & \secondbest{0.938} & \secondbest{0.987} & \secondbest{0.943} & \secondbest{0.936} & \secondbest{0.985} & \secondbest{0.925} & \secondbest{0.922} & \secondbest{0.954} & \secondbest{0.870} & \secondbest{0.820} & \best{0.898} \\
    \rowcolor{ourscolor_2}
    \textit{Samba+ (Ours)} & 59.03 & 72.50 & \best{0.826} & \best{0.737} & \best{0.856} & \best{0.955} & \best{0.949} & \best{0.991} & \best{0.947} & \best{0.939} & \best{0.986} & \best{0.926} & \best{0.922} & \best{0.956} & \best{0.873} & \best{0.823} & \secondbest{0.890} \\
    \bottomrule
  \end{tabular}
  \label{VSOD_SOTA}
  \vspace{-4mm}
\end{table*}

\begin{table*}[t]
  \definecolor{subheadercolor}{gray}{0.92}
  \definecolor{ourscolor_1}{HTML}{FEF7F2}
  \renewcommand{\arraystretch}{0.7}
  \setlength{\tabcolsep}{2.2pt}
  \centering
  \small
  \caption{Quantitative comparison of our \textit{Samba} and \textit{Samba+} against other SOTA RGB-T SOD methods on three benchmark datasets. 
  The best and second-best results are highlighted in \textcolor{red}{red} and \secondbest{blue}, respectively.}
  \vspace{-2mm}
  
    \begin{tabular}{lr|*{5}{>{\centering\arraybackslash}p{1.29cm}}|
                    *{4}{> {\centering\arraybackslash}p{1.29cm}}|
                    *{2}{>{\centering\arraybackslash}p{1.29cm}}}

      \toprule
      \multicolumn{2}{l|}{\multirow{3}{*}{Method}} &
      \multicolumn{5}{c|}{\cellcolor{subheadercolor}\textbf{\textit{CNN-based}}} &
      \multicolumn{4}{c|}{\cellcolor{subheadercolor}\textbf{\textit{Transformer-based}}} &
      \multicolumn{2}{c}{\cellcolor{subheadercolor}\textbf{\textit{Ours}}} \\

      \cmidrule(lr){3-7}\cmidrule(lr){8-11}\cmidrule(lr){12-13}
      
            && 
      \text{CGFNet}$_{21}$ & \text{MGAI}$_{22}$ & \text{TNet}$_{22}$ & \text{CGMDR}$_{22}$ & \text{SPNet}$_{23}$ & \text{SwinNet}$_{21}$
       & \text{HRTrans}$_{22}$ & \text{VSCode}$_{24}$ & \text{ConTri}$_{25}$ &
      \multirow{2}{*}{\textit{Samba}} &
      \multirow{2}{*}{\textit{Samba+}} \\
      
      &&
      \cite{wang2021cgfnet} &
      \cite{song2022multiple} &
      \cite{cong2022does} &
      \cite{chen2022cgmdrnet} &
      \cite{zhang2023saliency} &
      \cite{liu2021swinnet} &
      \cite{tang2022hrtransnet} &
      \cite{luo2024vscode} &
      \cite{Tangdivide} &
      & \\
      
      \midrule
      \multicolumn{2}{l|}{Params (M)} & 69.92 & 87.09 & 87.04 & - & 104.03 & 199.18 & 68.89 & 74.72 & 96.31 & 54.94 & 59.03 \\
      \multicolumn{2}{l|}{MACs (G)}   & 382.63 & 78.37 & 54.90 & - & 67.59 & 122.20 & 18.80 & 93.76 & 126.88 & 71.64 & 72.50 \\
      \midrule

      \multirow{3}{*}{VT821}
      & $S_m\uparrow$ & 0.881 & 0.891 & 0.899 & 0.894 & 0.913 & 0.904 & 0.906 & 0.926 & 0.916 & \secondbest{0.934} & \best{0.944} \\
      & $F_m\uparrow$ & 0.866 & 0.870 & 0.885 & 0.872 & 0.900 & 0.877 & 0.881 & 0.910 & 0.896 & \secondbest{0.927} & \best{0.940} \\
      \cite{wang2018rgb} & $E_m\uparrow$ & 0.920 & 0.933 & 0.936 & 0.932 & 0.949 & 0.937 & 0.944 & 0.954 & 0.948 & \secondbest{0.965} & \best{0.971} \\
      \midrule

      \multirow{3}{*}{VT5000}
      & $S_m\uparrow$ & 0.883 & 0.884 & 0.895 & 0.896 & 0.914 & 0.912 & 0.912 & 0.925 & 0.924 & \secondbest{0.928} & \best{0.932} \\
      & $F_m\uparrow$ & 0.852 & 0.846 & 0.864 & 0.877 & 0.905 & 0.885 & 0.895 & 0.900 & 0.918 & \secondbest{0.919} & \best{0.927} \\
      \cite{tu2022rgbt} & $E_m\uparrow$ & 0.926 & 0.930 & 0.936 & 0.939 & 0.954 & 0.944 & 0.948 & 0.959 & \secondbest{0.963} & \secondbest{0.963} & \best{0.976} \\
      \midrule

      \multirow{3}{*}{VT1000}
      & $S_m\uparrow$ & 0.923 & 0.929 & 0.929 & 0.931 & 0.941 & 0.938 & 0.938 & 0.952 & 0.941 & \secondbest{0.953} & \best{0.955} \\
      & $F_m\uparrow$ & 0.923 & 0.921 & 0.921 & 0.927 & 0.943 & 0.933 & 0.931 & \secondbest{0.947} & 0.939 & \best{0.956} & \best{0.956} \\
      \cite{tu2019rgb} & $E_m\uparrow$ & 0.959 & 0.965 & 0.965 & 0.966 & 0.975 & 0.974 & 0.969 & \secondbest{0.981} & 0.976 & \best{0.983} & \secondbest{0.981} \\

      \bottomrule
    \end{tabular}%
  \vspace{-4mm}
  \label{RGBT_SOTA}
\end{table*}


\newlength{\metlabw}
\settowidth{\metlabw}{$E_m$} 
\newcommand{\met}[1]{\makebox[\metlabw][r]{$#1_m$}\,$\uparrow$} 


\begin{table}[t]
  \centering
  \small
  \setlength{\tabcolsep}{3.2pt}
  \definecolor{ourscolor_1}{RGB}{245, 245, 245}
  \definecolor{ourscolor_2}{RGB}{230, 230, 230}
  \caption{Quantitative comparison of \textit{Samba} and \textit{Samba+} against other SOTA RGB-D VSOD methods on three public datasets. The best and second-best results are highlighted in \best{red} and \secondbest{blue}.}
  \vspace{-2mm}
  \renewcommand{\arraystretch}{0.7}
  \begin{tabular}{ll|ccc|c|c}
  \toprule
  \multirow{2}{*}{\raggedright Method} & 
  \multicolumn{1}{c|}{} &
  \multicolumn{1}{c}{DCTNet+} &
  \multicolumn{1}{c}{DVSOD} &
  \multicolumn{1}{c|}{ATFNet} &
  \multirow{2}{*}{\raggedright \textit{Samba}} &  
  \multirow{2}{*}{\raggedright \textit{Samba+}} \\ 

  & &
  \cite{mou2023salient}$_{24}$ &
  \cite{li2024dvsod}$_{24}$ &
  \cite{lin2024vidsod}$_{24}$ &
  & \\

  \midrule
  \multicolumn{2}{l|}{Params (M)} & 90.69 & 97.34 & 124.07 & 60.28 & 59.03 \\
  \multicolumn{2}{l|}{MACs (G)}  & 117.94 & 276.46 & 54.36 & 96.60 & 96.52 \\
  \midrule

  \multirow{3}{*}{\raggedright RDVS}
    & \met{S} & 0.869 & 0.689 & 0.741 & \secondbest{0.883} & \best{0.893} \\
    & \met{F} & 0.814 & 0.574 & 0.592 & \secondbest{0.834} & \best{0.840} \\
    & \met{E} & 0.914 & 0.733 & 0.785 & \best{0.936} & \secondbest{0.920} \\
  \midrule

  \multirow{3}{*}{\raggedright DVisal}
    & \met{S} & 0.814 & 0.729 & 0.723 & \secondbest{0.847} & \best{0.875} \\
    & \met{F} & 0.807 & 0.648 & 0.659 & \secondbest{0.825} & \best{0.854} \\
    & \met{E} & \secondbest{0.909} & 0.813 & 0.809 & \best{0.914} & \best{0.914} \\
  \midrule

  \multirow{3}{*}{\raggedright ViDSOD}
    & \met{S} & 0.877 & 0.770 & 0.875 & \secondbest{0.923} & \best{0.924} \\
    & \met{F} & 0.820 & 0.687 & 0.832 & \secondbest{0.895} & \best{0.913} \\
    & \met{E} & 0.901 & 0.846 & 0.911 & \secondbest{0.944} & \best{0.948} \\
  \bottomrule
  \end{tabular}

  \label{RGBDVSOD_SOTA}
  \vspace{-4mm}
\end{table}

\begin{table}[t]
  \definecolor{ourscolor_1}{HTML}{FEF7F2}
  \centering
  \small 
  \caption{Quantitative comparison of our \textit{Samba} and  \textit{Samba+}~against other SOTA methods on VDT-2048. The best and second-best results are highlighted in \best{red} and \secondbest{blue}, respectively.}
  \vspace{-2mm}
  \label{tab:vdt_sota}
  \renewcommand{\arraystretch}{0.83} 
  \setlength{\tabcolsep}{2.8pt} 

  \begin{tabular}{l|l|cc| ccc}
    \toprule
    \multirow{2.5}{*}{Method} & \multirow{2.5}{*}{Type} & \multirow{2.5}{*}{\makecell{Params \\ (M)}}  & \multirow{2.5}{*}{\makecell{MACs \\ (G)}}  & \multicolumn{3}{c}{VDT-2048\cite{Song2023HWSI}} \\
    \cmidrule(l){5-7}
     & & & & \textit{Fm}$\uparrow$ & \textit{Sm}$\uparrow$ & \textit{Em}$\uparrow$ \\
    \midrule
    \text{DPA}$_{20}$ \cite{Chen2020DPANet} & RGB-D & 92.39 & 29.48 & 0.459 & 0.818 & 0.686 \\
    \text{SwinNet}$_{21}$ \cite{liu2021swinnet} & RGB-D & 124.72 & 99.39 & 0.729 & 0.919 & 0.896 \\
    \text{CGFNet}$_{21}$ \cite{wang2021cgfnet} & RGB-T & 69.92 & 173.89 & 0.778 & 0.917 & 0.930 \\
    \text{LSNet}$_{23}$ \cite{Zhou2023LSNet} & RGB-T & 4.57 & 1.23 & 0.743 & 0.888 & 0.920 \\
    \text{CAFC}$_{24}$ \cite{Jin2024CAFC} & RGB-T & - & - & 0.854 & 0.913 & 0.977 \\
    \text{HWSI}$_{23}$ \cite{Song2023HWSI} & VDT & 124.72 & 178.95 & 0.872 & 0.931 & 0.981 \\
    \text{TMNet}$_{24}$ \cite{Wan2024TMNet} & VDT & - & 242.01 & 0.897 & 0.933 & 0.989 \\
    \text{MFFNet}$_{23}$ \cite{Wan2023MFFNet} & VDT & 103.24 & 539.80 & 0.901 & 0.936 & \secondbest{0.990} \\
    \text{DWFPR}$_{24}$ \cite{Luo2024DWFPRNet} & VDT & - & 253.35 & 0.901 & \secondbest{0.938} & \secondbest{0.990} \\
    \midrule
    \textit{Samba (Ours)} & VDT & 60.28 & 96.60 & \secondbest{0.910} & \secondbest{0.938} & \secondbest{0.990} \\
    \textit{Samba+ (Ours)} & Unified & 59.03 & 96.52 & \best{0.917} & \best{0.939} & \best{0.991} \\
    \bottomrule
  \end{tabular}
  \vspace{-6mm}
  \label{VDT_SOTA}
\end{table}

\subsection{Datasets and Metrics}
For \colorbox{gray!20}{RGB SOD}, we evaluate \textit{Samba} and \textit{Samba+} on five commonly used benchmark datasets, i.e., \textbf{DUTS} \cite{wang2017learning}, \textbf{DUT-O} \cite{yang2013saliency}, \textbf{HKU-IS} \cite{li2015visual}, \textbf{PASCAL-S} \cite{li2014secrets} and \textbf{ECSSD} \cite{yan2013hierarchical}. 
As for \colorbox{gray!20}{RGB-D SOD}, we use five benchmark datasets, including \textbf{NJUD} \cite{ju2014depth}, \textbf{NLPR} \cite{peng2014rgbd}, \textbf{SIP} \cite{fan2020rethinking}, \textbf{STERE} \cite{niu2012leveraging} and \textbf{DUTLF-D} \cite{piao2019depth}.
Regarding \colorbox{gray!20}{RGB-T SOD}, we employ three benchmark datasets: \textbf{VT821} \cite{wang2018rgb}, \textbf{VT1000} \cite{tu2019rgb} and \textbf{VT5000} \cite{tu2022rgbt}. 
For \colorbox{gray!20}{VSOD}, we utilize five benchmark datasets: \textbf{DAVIS} \cite{perazzi2016benchmark}, \textbf{DAVSOD-easy} \cite{fan2019shifting}, \textbf{FBMS} \cite{ochs2013segmentation}, \textbf{SegV2} \cite{li2013video} and \textbf{VOS} \cite{li2017benchmark}.
In terms of \colorbox{gray!20}{RGB-D VSOD}, three datasets are considered, including \textbf{RDVS} \cite{mou2023salient}, \textbf{DVisal} \cite{li2024dvsod} and \textbf{ViDSOD} \cite{lin2024vidsod}.
For \colorbox{gray!20}{VDT SOD}, we employ the \textbf{VDT-2048} \cite{Song2023HWSI} benchmark dataset.
We adopt three saliency metrics to evaluate model performance, i.e., structure-measure ($S_m$) \cite{fan2017structure}, maximum F-measure ($F_m$) \cite{achanta2009frequency} and maximum enhanced-alignment measure ($E_m$) \cite{fan2018enhanced}. To assess model computational complexity and model size, we report the multiply-accumulate operations (MACs) and parameters (Params).

\subsection{Implementation Details}

Our \textit{Samba} and \textit{Samba+} is implemented in PyTorch trained on four NVIDIA RTX 4090 GPUs. Following previous works, we have arranged the training sets for each task as follows: the training set (about 10.5k samples) of \textbf{DUTS} for \colorbox{gray!20}{RGB SOD}, the training sets (about 3k samples) of \textbf{NJUD}, \textbf{NLPR} and \textbf{DUTLF-D} for \colorbox{gray!20}{RGB-D SOD}, the training set (about 2.5k samples) of \textbf{VT5000} for \colorbox{gray!20}{RGB-T SOD}, the training sets (about 10.5k samples) of \textbf{DAVIS} and \textbf{DAVSOD} for \colorbox{gray!20}{VSOD}, the training set  (about 1k samples) of \textbf{VDT-2048} for \colorbox{gray!20}{VDT SOD}. Due to the lack of a benchmark training set for \colorbox{gray!20}{RGB-D VSOD}, we train and test \ourmodel~on \textbf{RDVS}, \textbf{DVisal} and \textbf{ViDSOD} separately. For \textit{Samba+}, we perform joint learning across aforementioned datasets  (about 6.8k samples). In preparing the training data, we found that a subset of samples appeared in both DVisal and ViDSOD. To avoid data leakage and ensure fair training, we only retain the samples from DVisal together with their ground-truth annotations. The final list of removed samples is released with our code. In the training process, we adopt AdamW optimizer with an initial learning rate of $1e-4$ and the batch size of 2. All input images are uniformly resized to 448 × 448 for training and testing, and are also augmented using various strategies like random flipping, random cropping and random rotating. Both models reach convergence within 40 training epochs, with metrics stabilizing thereafter.

Regarding computational complexity, the Params (or MACs) of \textit{Samba} and \textit{Samba+} models reach 60.28M (96.60G) and 59.03M (96.52G), respectively. 
The Params (or MACs) of each component are as follows: (1) SGMB 3.56M (1.91G), (2) CAU 5.87M (3.58G), (3) HGA 6.49M (1.27G), (4) MFM 10.69M (1.86G), and (5) SIR 0.27M (0.04G). These results indicate that the proposed modules are lightweight and efficient, having limited computational overhead. 

\subsection{Comparison with State-of-the-Art Methods}
\noindent
\textbf{Quantitative Evaluation.} 
Since our \textit{Samba} and \textit{Samba+} are unified models to handle all SOD tasks, we present comparative experiments against existing SOTA methods across six common SOD tasks, including 10 models for RGB SOD, 10 models for RGB-D SOD, 10 models for VSOD, 9 models for RGB-T SOD, 9 models for VDT SOD, 3 models for RGB-D VSOD, as show in Table~\ref{RGB_SOTA}, \ref{RGBD_SOTA}, \ref{VSOD_SOTA}, \ref{RGBT_SOTA}, \ref{RGBDVSOD_SOTA} and \ref{VDT_SOTA}. The comprehensive results illustrate that \textit{Samba} outperforms existing SOTA CNN- and transformer-based SOD models across 22 datasets, with a comparable number of Params and relatively low MACs, demonstrating the superior performance of \textit{Samba}. 
Building upon \textit{Samba}, the enhanced \textit{Samba+} further improves performance across these tasks and datasets by using a single trained versatile mode, while using fewer parameters and lower computational cost compared with \textit{Samba}’s tri-modal configuration. Specifically, for RGB SOD, both \textit{Samba} and \textit{Samba+} exhibit lower Params and MACs than those of transformer-based methods (except VST-S++~\cite{liu2024vst++}), while achieving best results on the given datasets. Although several CNN-based methods (e.g., ICON-R \cite{zhuge2022salient}, EDN \cite{wu2022edn} and CSF-R2 \cite{gao2020highly}) are more lightweight, their performance is substantially inferior to our frameworks. Regarding RGB-D/T SOD, VSOD, RGB-D VSOD and VDT SOD, most methods (except BBSNet \cite{fan2020bbs}, VST-S++ \cite{liu2024vst++}, FSNet \cite{ji2021full}, SPNet \cite{zhang2023saliency}, TNet \cite{cong2022does}, ATFNet \cite{lin2024vidsod}, LSNet \cite{Zhou2023LSNet}) exhibit higher computational complexity than \textit{Samba} and \textit{Samba+}, regardless of whether CNN- or transformer-based. Nevertheless, \textit{Samba} and \textit{Samba+} still outperforms these methods across all datasets.

Notably, although VSCode \cite{luo2024vscode} is jointly trained across diverse SOD tasks, its task-specific prompt offsets injected into the encoder and decoder layers make it effectively equivalent to maintaining separate parameter sets for each sub-task, limiting cross-task knowledge sharing. In contrast, our \textit{Samba+} adopts a fully shared architecture without task-specific components, building a more unified and scalable SOD framework.

\begin{figure*}[t!]
    \centering
    \captionsetup{skip=5pt}
    \resizebox{\textwidth}{!}{%
    \includegraphics[width=0.95\linewidth]{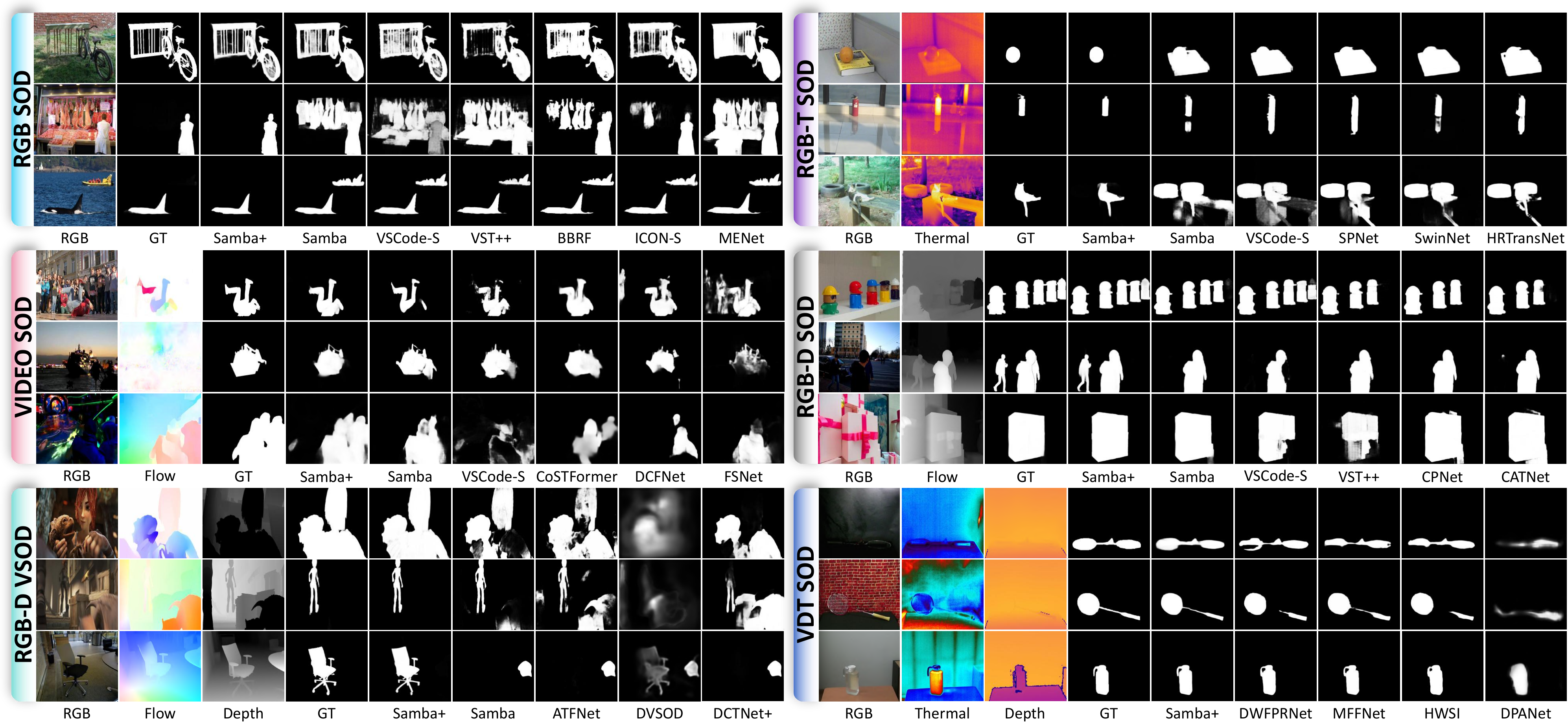}}
    \caption{Visual comparison against SOTA SOD methods.}
    \label{fig_visual}
    \vspace{-0.3cm}
\end{figure*}

\noindent
\textbf{Qualitative Evaluation.}
To qualitatively assess the effectiveness of \textit{Samba} and \textit{Samba+}, Fig.~\ref{fig_visual} presents visual comparisons with representative SOTA models across diverse SOD tasks. The selected examples include challenging cases such as hollow structures, multiple salient objects, complex backgrounds, and fine-grained saliency variations. In addition, from a multi-modal perspective, we provide instances where the RGB modality is degraded and complementary cues from flow, depth, or thermal modalities become essential. As illustrated in Fig.~\ref{fig_visual}, \textit{Samba} and \textit{Samba+} integrate cross-modal cues effectively, yielding accurate object localization and well-preserved boundaries, which reflects their strong applicability across diverse scenarios.

\begin{table*}[t]
  \centering
  \footnotesize 
  
  \caption{Ablation study of \textit{Samba} and \textit{Samba+} on three RGB SOD, three RGB-D SOD, and one RGB-D VSOD datasets. The best results are highlighted in \textbf{bold}. The settings are grouped by the ablated component: (A) SGMB, (B) CAU, (C) HGA, and (D) SIR .}
  \vspace{-2mm}
  \definecolor{ourscolor}{gray}{0.92} 
  \definecolor{color_red}{RGB}{252,182,165} 
  \definecolor{color_pink}{RGB}{255,217,178} 
  \definecolor{color_yellow}{RGB}{255,255,204} 
  \definecolor{color_blue1}{RGB}{225, 232, 250}
  \sisetup{detect-weight=true, detect-family=true}
  \setlength\tabcolsep{4pt} 
  \renewcommand{\arraystretch}{1.2} 
  \renewcommand{\tabcolsep}{1.1mm} 
  \resizebox{\textwidth}{!}{%
  \begin{tabular}{l *{7}{S[table-format=1.3] S[table-format=1.3] S[table-format=1.3]}}
    
    \toprule 
    \multirow{3.5}{*}{Settings} & \multicolumn{9}{c}{RGB SOD} & \multicolumn{9}{c}{RGB-D SOD} & \multicolumn{3}{c}{RGB-D VSOD} \\
    \cmidrule(lr){2-10} \cmidrule(lr){11-19} \cmidrule(lr){20-22} 
    & \multicolumn{3}{c}{DUTS\cite{wang2017learning}} & \multicolumn{3}{c}{ECSSD\cite{yan2013hierarchical}} & \multicolumn{3}{c}{DUT-O\cite{yang2013saliency}} & \multicolumn{3}{c}{NJUD\cite{ju2014depth}} & \multicolumn{3}{c}{NLPR\cite{peng2014rgbd}} & \multicolumn{3}{c}{DUTLF-D\cite{piao2019depth}} & \multicolumn{3}{c}{RDVS\cite{mou2023salient}} \\
    \cmidrule(r){2-4} \cmidrule(r){5-7} \cmidrule(r){8-10} \cmidrule(r){11-13} \cmidrule(r){14-16} \cmidrule(r){17-19} \cmidrule(r){20-22}
    & \metricsheader & \metricsheader & \metricsheader & \metricsheader & \metricsheader & \metricsheader & \metricsheader \\
    \midrule 

 
    \textit{\textbf{Samba+ (Unified)}}
    & \bfseries 0.936 & \bfseries 0.932 & \bfseries 0.969 & \bfseries 0.902 & \bfseries 0.875 & \bfseries 0.934 & \bfseries 0.955 & \bfseries 0.966 & \bfseries 0.978
    & \bfseries 0.950 & \bfseries 0.958 & \bfseries 0.979 & \bfseries 0.950 & \bfseries 0.944 & \bfseries 0.977 & \bfseries 0.960 & \bfseries 0.967 & \bfseries 0.979
    & \bfseries 0.888 & \bfseries 0.838 & \bfseries 0.939 \\
    \textit{\textbf{Samba (Task-Specific)}}
    & \bfseries 0.932 & \bfseries 0.930 & \bfseries 0.966 & \bfseries 0.889 & \bfseries 0.859 & \bfseries 0.922 & \bfseries 0.953 & \bfseries 0.965 & \bfseries 0.978
    & \bfseries 0.949 & \bfseries 0.956 & \bfseries 0.975 & \bfseries 0.947 & \bfseries 0.941 & \bfseries 0.976 & \bfseries 0.956 & \bfseries 0.964 & \bfseries 0.976
    & \bfseries 0.883 & \bfseries 0.834 & \bfseries 0.936 \\
    \midrule 

    \rowcolor{color_red}
    \multicolumn{22}{l}{\textit{A: Ablation on \textbf{SGMB} (Saliency-guided mamba block)}} \\
    \addlinespace[0.7ex]  
    A1 (w/o SGMB)         & 0.922 & 0.916 & 0.950 & 0.881 & 0.843 & 0.909 & 0.946 & 0.954 & 0.966 & 0.940 & 0.941 & 0.961 & 0.942 & 0.932 & 0.965 & 0.947 & 0.952 & 0.963 & 0.876 & 0.822 & 0.929 \\
    A2 (Replace $\text{SG-SS2D} \rightarrow \text{SS2D}$)  & 0.926 & 0.919 & 0.952 & 0.882 & 0.846 & 0.914 & 0.946 & 0.952 & 0.964 & 0.941 & 0.945 & 0.966 & 0.945 & 0.933 & 0.962 & 0.949 & 0.957 & 0.968 & 0.878 & 0.827 & 0.930 \\
    A3 (Scan - Fig.~\ref{fig_intro} (a))    & 0.929 & 0.924 & 0.961 & 0.884 & 0.852 & 0.917 & 0.950 & 0.961 & 0.973 & 0.945 & 0.949 & 0.970 & 0.944 & 0.937 & 0.974 & 0.949 & 0.959 & 0.971 & 0.880 & 0.829 & 0.931 \\
    A4 (Scan - Fig.~\ref{fig_intro} (b))    & 0.928 & 0.927 & 0.963 & 0.886 & 0.853 & 0.917 & 0.947 & 0.960 & 0.971 & 0.943 & 0.948 & 0.965 & 0.943 & 0.938 & 0.972 & 0.947 & 0.956 & 0.972 & 0.879 & 0.831 & 0.932 \\
    A5 (Scan - Fig.~\ref{fig_intro} (c))    & 0.930 & 0.928 & 0.963 & 0.886 & 0.857 & 0.918 & 0.951 & 0.962 & 0.973 & 0.946 & 0.952 & 0.972 & 0.945 & 0.940 & 0.973 & 0.949 & 0.961 & 0.972 & 0.881 & 0.831 & 0.933 \\
    A6 (w/o path direction variants)      & 0.928 & 0.926 & 0.963 & 0.886 & 0.856 & 0.918 & 0.948 & 0.960 & 0.971 & 0.947 & 0.954 & 0.973 & 0.945 & 0.940 & 0.972 & 0.952 & 0.962 & 0.975 & 0.880 & 0.830 & 0.933 \\
    \midrule 

    \rowcolor{color_pink}
    \multicolumn{22}{l}{\textit{B: Ablation on \textbf{CAU} (Context-aware upsampling)}} \\
    \addlinespace[0.7ex] 
    B1 (Nearest-neighbor) & 0.927 & 0.923 & 0.960 & 0.887 & 0.856 & 0.912 & 0.952 & 0.961 & 0.976 & 0.941 & 0.952 & 0.966 & 0.942 & 0.939 & 0.972 & 0.953 & 0.957 & 0.969 & 0.879 & 0.828 & 0.932 \\
    B2 (DUpsampling)      & 0.921 & 0.918 & 0.952 & 0.877 & 0.848 & 0.905 & 0.949 & 0.955 & 0.970 & 0.936 & 0.944 & 0.962 & 0.939 & 0.933 & 0.965 & 0.947 & 0.954 & 0.961 & 0.874 & 0.819 & 0.928 \\
    B3 (DySample)         & 0.923 & 0.924 & 0.956 & 0.884 & 0.855 & 0.913 & 0.946 & 0.959 & 0.968 & 0.937 & 0.946 & 0.966 & 0.938 & 0.933 & 0.963 & 0.950 & 0.955 & 0.966 & 0.878 & 0.826 & 0.933 \\
    B4 (CAU - 1 patch)    & 0.931 & 0.928 & 0.964 & 0.887 & 0.857 & 0.921 & 0.951 & 0.965 & 0.974 & 0.947 & 0.955 & 0.972 & 0.946 & 0.939 & 0.971 & 0.955 & 0.963 & 0.974 & 0.882 & 0.834 & 0.934 \\
    B5 (CAU - 3 patches)  & 0.929 & 0.926 & 0.964 & 0.888 & 0.854 & 0.920 & 0.950 & 0.964 & 0.976 & 0.948 & 0.953 & 0.971 & 0.944 & 0.936 & 0.973 & 0.955 & 0.962 & 0.971 & 0.879 & 0.832 & 0.931 \\
    B6 (CAU - 5 patches)  & 0.926 & 0.922 & 0.963 & 0.888 & 0.853 & 0.916 & 0.949 & 0.961 & 0.974 & 0.945 & 0.952 & 0.971 & 0.940 & 0.934 & 0.972 & 0.952 & 0.959 & 0.972 & 0.877 & 0.833 & 0.928 \\
    \midrule 

    \rowcolor{color_yellow}
    \multicolumn{22}{l}{\textit{C: Ablation on \textbf{HGA} (Hub-and-spoke graph attention fusion module)}} \\
    \addlinespace[0.7ex] 
    Samba+GAM          & 0.933 & 0.931 & 0.967 & 0.900 & 0.873 & 0.933 & 0.954 & 0.965 & 0.978 & 0.949 & 0.957 & 0.977 & 0.949 & 0.943 & 0.977 & 0.958 & 0.966 & 0.978 & 0.886 & 0.836 & 0.937 \\
    C1 (MFM - Samba task-specific)     & 0.932 & 0.930 & 0.966 & 0.889 & 0.859 & 0.922 & 0.953 & 0.965 & 0.978 & 0.949 & 0.956 & 0.975 & 0.947 & 0.941 & 0.976 & 0.956 & 0.964 & 0.976 & 0.883 & 0.834 & 0.936 \\
    C2 (RFM)              & 0.929 & 0.926 & 0.965 & 0.886 & 0.857 & 0.918 & 0.951 & 0.962 & 0.973 & 0.946 & 0.953 & 0.973 & 0.944 & 0.938 & 0.972 & 0.953 & 0.963 & 0.972 & 0.881 & 0.833 & 0.932 \\
    C3 (Concat + Conv)      & 0.932 & 0.930 & 0.966 & 0.889 & 0.859 & 0.922 & 0.953 & 0.965 & 0.978 & 0.944 & 0.954 & 0.970 & 0.945 & 0.936 & 0.973 & 0.952 & 0.961 & 0.969 & 0.878 & 0.831 & 0.926 \\
    C4 (Add + Pool)      & 0.928 & 0.924 & 0.963 & 0.883 & 0.854 & 0.915 & 0.947 & 0.961 & 0.971 & 0.944 & 0.953 & 0.968 & 0.943 & 0.934 & 0.971 & 0.951 & 0.959 & 0.967 & 0.876 & 0.829 & 0.925 \\
    \midrule

    \rowcolor{color_blue1}
    \multicolumn{22}{l}{\textit{D: Ablation on \textbf{SIR} (Synergistic integrity refinement module)}} \\
    \addlinespace[0.7ex] 
    Samba+SIR    & 0.933 & 0.932 & 0.968 & 0.893 & 0.863 & 0.923 & 0.952 & 0.965 & 0.978 & 0.949 & 0.958 & 0.976 & 0.948 & 0.942 & 0.976 & 0.958 & 0.965 & 0.972 & 0.885 & 0.836 & 0.937 \\
    \bottomrule 
  \end{tabular}}
  \label{ablation}
  \vspace{-0.3cm}
\end{table*}

 \subsection{Ablation Study}
 \label{sec:ablation}
To verify relative contributions of components in \textit{Samba} and \textit{Samba+}, we conduct thorough ablation studies by removing or replacing them from our models. As shown in Table \ref{ablation}, we perform a range of experiments on three \colorbox{gray!20}{RGB SOD} (DUTS, ECSSD, DUT-O), three \colorbox{gray!20}{RGB-D SOD} (NJUD, NLPR, DUTLF-D) and one \colorbox{gray!20}{RGB-D VSOD} datasets (RDVS). 

\noindent
\textbf{Effectiveness of SGMB.} 
To validate the effectiveness of SGMB, we first delete the SGMB module, denoted as variant ``A1'' in Table~\ref{ablation}, and then utilize an SS2D module to replace the SG-SS2D module of SGMB, denoted as variant ``A2''. From the comparison results, it is evident that our model performs better than ``A1'' and ``A2'', which indicates the significant contribution of SGMB in boosting detection performance. Since the core idea of SGMB lies in the SNS algorithm, which is used to maintain spatial continuity of salient patches. To evaluate its contribution, we apply three other scanning manners to salient regions, i.e., Fig.~\ref{fig_intro} (a), (b) and (c), which fail to preserve spatial continuity of salient patches. We denote these evaluations as ``A3'', ``A4'' and ``A5''. Compared to these variants, our model consistently outperforms them on the evaluated datasets, highlighting the importance of preserving spatial continuity of salient patches. In the SNS algorithm, we generate three variants of the scanning path by changing the directions to enhance the robustness of SNS. To validate the contribution of the three paths, we create three copies of the initial scanning path to replace them, denoted as ``A6''. The comparison results between ``Ours'' and ``A6'' demonstrate the effectiveness of using multiple paths with varying directions.

\noindent
\textbf{Effectiveness of CAU.}
The proposed CAU module achieves learnable upsampling by modeling contextual dependencies between hierarchical features, which facilitates the alignment and aggregations of these features. To evaluate its effectiveness, we compare it with three other upsampling methods. Firstly, we utilize nearest-neighbor interpolation to replace the upsampling operation in the CAU module, denoted as variant ``B1''. Besides, we investigate two other learnable upsampling methods: DUpsampling \cite{tian2019decoders} and DySample \cite{liu2023learningto}, and apply them separately for feature upsampling, denoted as ``B2'' and ``B3''.  Comparing ``B1'', ``B2'' and ``B3'' with our full model, it is clear that CAU outperforms all alternative upsampling methods. In the CAU module, we introduce a novel patch pairing and ordering scheme, which plays a central role in the upsampling process. To validate the effectiveness of this design, we shift the original pairing sequences by one patch, three patches, and five patches, resulting in variants ``B4'', ``B5'' and ``B6''. This observation suggests that dense-prediction upsampling requires both resolution recovery and preservation of local relational order, which detail conventional methods often overlook.

\noindent\textbf{Effectiveness of Converter.}
To comprehensively validate the effectiveness of the proposed HGA module, we conduct a series of substitution experiments. First, we replace HGA with \textit{Samba}'s task-specific converter (MFM), denoted as ``C1'' and re-implement the RFM module from \cite{mou2023salient} as ``C2''. These two variants serve as representative specialized cross-modal fusion strategies. In addition, to compare HGA with generic fusion baselines, we further substitute it with a simple feature concatenation followed by a \(1\times1\) convolution, and a modality-wise pooling operation that computes the mean feature map across all modalities, denoted as ``C3'' and ``C4''. The comparative results consistently demonstrate that our HGA module outperforms these alternatives. This performance gain primarily arises from HGA’s modality-agnostic formulation, which accommodates arbitrary modality combinations while jointly exploiting their shared and complementary cues.

\noindent\textbf{Effectiveness of SIR.}
To validate the effectiveness of SIR module, we integrate it into \textit{Samba} and compare it with the baseline devoid of SIR.  Table~\ref{ablation} demonstrates that introducing SIR consistently boosts performance across SOD datasets. This improvement stems from the capacity of SIR to restore the spatial integrity of intermediate features by rectifying the incomplete SNS scanning path.

\noindent\textbf{Effectiveness of Encoder Configuration.}
To verify the effectiveness of our selected encoder configuration, VMamba-S (D2), we conduct an ablation study by replacing it with its variants, VMamba-T (D1) and VMamba-B (D3). As shown in Table~\ref{Encoder_Ablation}, upgrading from D1 to D2 yields substantial performance gains, particularly on PASCAL-S, at the cost of a relatively modest increase in model complexity (+16.11M Parameters and +12.65G MACs). In contrast, further scaling from D2 to D3 leads to a significant surge in parameters (+38.32M) and MACs (+35.74G) with diminishing returns in performance. Consequently, VMamba-S (D2) achieves the optimal trade-off between efficiency and representational capacity and is adopted as our encoder. Surprisingly, this finding reveals that in structure-driven SOD tasks, scaling up Mamba-based models does not necessarily lead to performance gains.

\begin{table*}[t]
  \centering
  \footnotesize
  
  \definecolor{ourscolor}{HTML}{EAF6F8} 
  \sisetup{detect-weight=true, detect-family=true}
  \newcommand{\mcolc}[1]{\multicolumn{1}{c}{#1}} 

\setlength\tabcolsep{2.2pt} 
  \renewcommand{\arraystretch}{1.2} 
  \renewcommand{\tabcolsep}{1.2mm} 

  \caption{Ablation studies of different encoder configurations on five RGB benchmark datasets. $M$ represents mean absolute error (MAE)\cite{borji2015salient,perazzi2012saliency}. The best results are stressed in \textbf{bold}.}
  \vspace{-2mm}
  \resizebox{\textwidth}{!}{%
    \begin{tabular}{ll S[table-format=2.2] S[table-format=2.2] *{5}{S[table-format=.3] S[table-format=.3] S[table-format=.3] S[table-format=.3]}}
      \toprule
      \multirow{2}{*}{Variant} & \multirow{2}{*}{Backbone} & 
      \mcolc{\multirow{2}{*}{\vspace{-6pt}\makecell{Params\\(M)}}} & 
      \mcolc{\multirow{2}{*}{\vspace{-6pt}\makecell{MACs\\(G)}}} & 
      \multicolumn{4}{c}{DUTS\cite{wang2017learning}} & 
      \multicolumn{4}{c}{DUT-O\cite{yang2013saliency}} & 
      \multicolumn{4}{c}{HKU-IS\cite{li2015visual}} & 
      \multicolumn{4}{c}{PASCAL-S\cite{li2014secrets}} & 
      \multicolumn{4}{c}{ECSSD\cite{yan2013hierarchical}} \\
      \cmidrule(lr){5-8} \cmidrule(lr){9-12} \cmidrule(lr){13-16} \cmidrule(lr){17-20} \cmidrule(lr){21-24}
      {} & {} & \mcolc{} & \mcolc{} & 
      \mcolc{$S_m\uparrow$} & \mcolc{$F_m\uparrow$} & \mcolc{$E_m\uparrow$} & \mcolc{$M\downarrow$} &
      \mcolc{$S_m\uparrow$} & \mcolc{$F_m\uparrow$} & \mcolc{$E_m\uparrow$} & \mcolc{$M\downarrow$} &
      \mcolc{$S_m\uparrow$} & \mcolc{$F_m\uparrow$} & \mcolc{$E_m\uparrow$} & \mcolc{$M\downarrow$} &
      \mcolc{$S_m\uparrow$} & \mcolc{$F_m\uparrow$} & \mcolc{$E_m\uparrow$} & \mcolc{$M\downarrow$} &
      \mcolc{$S_m\uparrow$} & \mcolc{$F_m\uparrow$} & \mcolc{$E_m\uparrow$} & \mcolc{$M\downarrow$}\\
      \midrule
      D1 & VMamba-T & 33.48 & 34.03 & .925 & .921 & .958 & .023 & .881 & .847 & .914 & .041 & .939 & .952 & .971 & .021 & .883 & .885 & .921 & .052 & .949 & .961 & .972 & .022 \\
      \rowcolor{ourscolor}
      D2 & VMamba-S & 49.59 & 46.68 & .932 & .930 & .966 & \bfseries .020 & .889 & \bfseries .859 & .922 & \bfseries .037 & .945 & \bfseries .956 & .978 & \bfseries .018 & .892 & .896 & .931 & \bfseries .047 & .953 & .965 & .978 & .019 \\
      D3 & VMamba-B & 87.91 & 82.42 & \bfseries .934 & \bfseries .933 & \bfseries .967 & \bfseries .020 & \bfseries .890 & .857 & \bfseries .926 & \bfseries .037 & \bfseries .947 & \bfseries .956 & \bfseries .979 & \bfseries .018 & \bfseries .896 & \bfseries .899 & \bfseries .935 & .048 & \bfseries .956 & \bfseries .967 & \bfseries .979 & \bfseries .018 \\
      \bottomrule
    \end{tabular}
  }
  \vspace{-2mm}
  \label{Encoder_Ablation}
\end{table*}


\begin{table*}[t]
\centering
\footnotesize 

\caption{Ablation study on the continual learning strategies for \textit{Samba+}. We compare our MACL against three baselines. Training time is benchmarked on 4$\times$ NVIDIA RTX 4090 GPUs. The best results are highlighted in \textbf{bold}.}
\label{tab:ablation_cl_strategy} 

\definecolor{ourscolor}{HTML}{EAF6F8} 
\vspace{-2mm}
\sisetup{detect-weight=true, detect-family=true}
\setlength\tabcolsep{2.2pt} 
\renewcommand{\arraystretch}{1.2} 
\renewcommand{\tabcolsep}{1.2mm} 

\resizebox{\textwidth}{!}{%
\begin{tabular}{l c *{8}{S[table-format=1.3] S[table-format=1.3] S[table-format=1.3]}}
\toprule

\multirow{3}{*}{\vspace{-8pt}Strategy}

& \multirow{3}{*}{\vspace{-8pt}\makecell{Time\\(min)}} 
& \multicolumn{6}{c}{RGB SOD} 
& \multicolumn{6}{c}{RGB-D SOD} 
& \multicolumn{3}{c}{VSOD} 
& \multicolumn{3}{c}{RGB-T} 
& \multicolumn{3}{c}{RGB-D VSOD} 
& \multicolumn{3}{c}{VDT} \\
\cmidrule(lr){3-8} \cmidrule(lr){9-14} \cmidrule(lr){15-17} \cmidrule(lr){18-20} \cmidrule(lr){21-23} \cmidrule(lr){24-26}

& & \multicolumn{3}{c}{DUTS\cite{wang2017learning}} & \multicolumn{3}{c}{DUT-O\cite{yang2013saliency}} & \multicolumn{3}{c}{SIP\cite{fan2020rethinking}} & \multicolumn{3}{c}{NLPR\cite{peng2014rgbd}} & \multicolumn{3}{c}{SegV2\cite{li2013video}} & \multicolumn{3}{c}{VT5000\cite{tu2022rgbt}} & \multicolumn{3}{c}{RDVS\cite{mou2023salient}} & \multicolumn{3}{c}{VDT-2048\cite{Song2023HWSI}} \\
\cmidrule(lr){3-5} \cmidrule(lr){6-8} \cmidrule(lr){9-11} \cmidrule(lr){12-14} \cmidrule(lr){15-17} \cmidrule(lr){18-20} \cmidrule(lr){21-23} \cmidrule(lr){24-26}

& & {$S_m\uparrow$} & {$F_m\uparrow$} & {$E_m\uparrow$} & {$S_m\uparrow$} & {$F_m\uparrow$} & {$E_m\uparrow$} & {$S_m\uparrow$} & {$F_m\uparrow$} & {$E_m\uparrow$} & {$S_m\uparrow$} & {$F_m\uparrow$} & {$E_m\uparrow$} & {$S_m\uparrow$} & {$F_m\uparrow$} & {$E_m\uparrow$} & {$S_m\uparrow$} & {$F_m\uparrow$} & {$E_m\uparrow$} & {$S_m\uparrow$} & {$F_m\uparrow$} & {$E_m\uparrow$} & {$S_m\uparrow$} & {$F_m\uparrow$} & {$E_m\uparrow$} \\
\midrule
\addlinespace[2pt] 
Task-specific train. & 460 & 0.932 & 0.930 & 0.966 & 0.889 & 0.859 & 0.922 & 0.931 & 0.949 & 0.966 & 0.947 & 0.941 & \bfseries 0.976 & 0.943 & 0.938 & 0.987 & 0.928 & 0.919 & 0.963 & 0.883 & 0.834 & \bfseries 0.936 & 0.938 & 0.910 & 0.990 \\

Mixed training& 350 & 0.934 & 0.931 & 0.966 & 0.891 & 0.859 & \bfseries 0.927 & 0.934 & 0.953 & 0.969 & 0.939 & 0.934 & 0.970 & 0.953 & 0.946 & 0.990 & 0.930 & 0.924 & 0.966 & 0.891 & 0.837 & 0.933 & 0.924 & 0.895 & 0.985 \\
Sequential training& 280 & 0.931 & 0.930 & 0.966 & 0.891 & 0.856 & 0.926 & 0.938 & 0.954 & 0.971 & 0.941 & 0.935 & 0.971 & 0.951 & 0.944 & 0.989 & 0.931 & 0.926 & 0.967 & 0.892 & 0.839 & 0.920 & 0.932 & 0.910 & 0.990 \\
\rowcolor{ourscolor}
MACL (Ours) & 300 & \bfseries 0.936 & \bfseries 0.933 & \bfseries 0.968 & \bfseries 0.893 & \bfseries 0.860 & \bfseries 0.927 & \bfseries 0.948 & \bfseries 0.960 & \bfseries 0.977 & \bfseries 0.949 & \bfseries 0.944 & 0.975 & \bfseries 0.955 & \bfseries 0.949 & \bfseries 0.991 & \bfseries 0.932 & \bfseries 0.927 & \bfseries 0.976 & \bfseries 0.893 & \bfseries 0.840 & 0.920 & \bfseries 0.939 & \bfseries 0.917 & \bfseries 0.991 \\
\bottomrule
\end{tabular}}
\label{table_MACL}
\vspace{-2mm}
\end{table*}

\begin{figure*}[t!]
    \centering
    \captionsetup{skip=5pt}
    \includegraphics[width=1\linewidth]{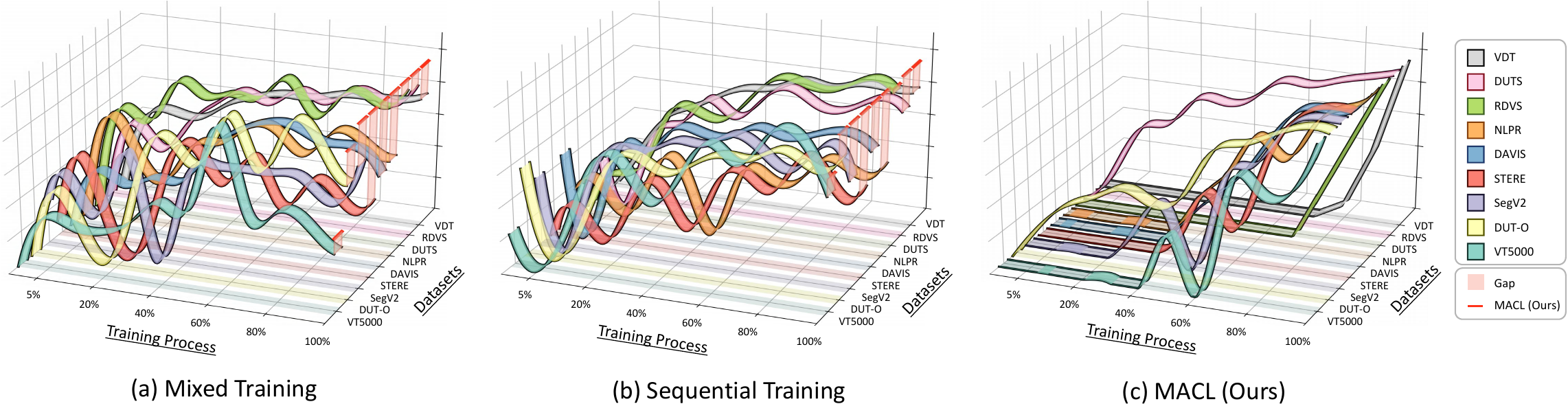}
    \caption{Visualization of the performance evolution under three training strategies.}
    \label{fig_ab}
    \vspace{-4mm}
\end{figure*}

\subsection{Analyses of Multi-Modal Joint Training in SOD}
\label{sec:joint training}
To analyze multi-modal joint training in SOD and validate our MACL strategy, we compare our method against three representative joint-training paradigms. Specifically, the comparison covers single-modal, dual-modal, and tri-modal SOD settings, as detailed in Table~\ref{table_MACL}.
\colorbox{gray!20}{(1)~Task-specific training} achieve a performance upper bound but suffer from practical limitations. They necessitate maintaining distinct architectures for each modality, leading to a massive parameter count (164.81M) and distinct training sessions that result in a prolonged total training duration (460 minutes).
\colorbox{gray!20}{(2)~Mixed joint training}, which creates a batch from all tasks, eliminates the need for architectural switching. However, it yields suboptimal performance and requires substantial training time (350 minutes), primarily due to severe convergence difficulties arising from modality interference and data imbalance.
\colorbox{gray!20}{(3)~Sequential joint training} introduces datasets from different modalities in a sequence. While it achieves the shortest training duration (280 minutes), it incurs significant performance degradation. This deficit is largely attributed to catastrophic forgetting and inter-modal conflicts, preventing the model from retaining knowledge across sequential tasks.
In contrast, MACL achieves a superior trade-off between efficiency and performance via its three-stage design. The anchoring stage stabilizes learning on primary RGB datasets, while the progressive adaptation stage effectively mitigates inter-modal conflicts.

To further investigate the training dynamics, Fig.~\ref{fig_ab} visualizes the performance evolution of strategies. The plot spans training progress (X-axis), dataset categories (Y-axis), and the normalized $S_m$ metric (Z-axis). The mixed joint training strategy exhibits strong oscillations—performance on VT5000 and DUT-O drops sharply after intermediate peaks—indicating unresolved cross-modal conflicts. Similarly, sequential training also fails to maintain stability, showing a substantial performance gap compared to the task-specific models. 
These observations demonstrate that MACL mitigates the instability and forgetting of naive joint training while enabling stable knowledge accumulation.

\noindent
\textbf{Some Insights.} As one of the earliest studies to incorporate continual learning into the Mamba architecture \cite{cheng2024mamba,zhao2024learning}, our study uncovered a counterintuitive phenomenon: jointly training the Mamba-based model on multi-modal and multi-task SOD datasets may fail to yield consistent improvements on individual datasets. In some datasets shown in Table~\ref{table_MACL}, joint training even led to performance degradation. This challenges the classical intuition and long-standing consensus that more training data naturally improves model generalization. Our analysis reveals that although all SOD tasks aim to detect salient objects, the underlying heterogeneous distributions across modalities and tasks can intensify optimization conflicts during training. From this standpoint, multi-modal and multi-task joint training resembles a gradual, stepwise alignment of representations rather than a synchronized optimization process.


\begin{table*}[t]
  \definecolor{subheadercolor}{gray}{0.92} 
  

  \definecolor{ourscolor}{HTML}{FEF7F2} 
  \centering
  \small 
  \caption{Results on camouflaged object detection (COD). The best and second-best results are highlighted in \best{red} and \secondbest{blue}, respectively.}
  \vspace{-2mm}
  \renewcommand{\arraystretch}{0.8} 
  \setlength\tabcolsep{2pt} 

  \begin{tabular}{l|l|cccc|cccc|cccc|cccc}
    \toprule
    \multirow{2.5}{*}{Method} & \multirow{2.5}{*}{Venue} & \multicolumn{4}{c}{CHAMELEON\!\cite{Skurowski2018Chameleon}} & \multicolumn{4}{c}{COD10K\!\cite{Fan2020SINet}} & \multicolumn{4}{c}{CAMO\!\cite{Le2019Anabranch}} & \multicolumn{4}{c}{NC4K\!\cite{Lv2021LSR}} \\
    \cmidrule(lr){3-6} \cmidrule(lr){7-10} \cmidrule(lr){11-14} \cmidrule(lr){15-18}
    & & $S_m\uparrow$ & $M\downarrow$ & $E_\phi\uparrow$ & $F_\beta\uparrow$ & $S_m\uparrow$ & $M\downarrow$ & $E_\phi\uparrow$ & $F_\beta\uparrow$ & $S_m\uparrow$ & $M\downarrow$ & $E_\phi\uparrow$ & $F_\beta\uparrow$ & $S_m\uparrow$ & $M\downarrow$ & $E_\phi\uparrow$ & $F_\beta\uparrow$ \\
    \midrule
    \multicolumn{18}{c}{\cellcolor{subheadercolor}\textbf{\textit{CNN-based Methods}}} \\
    \midrule
    FEDER\cite{He2023FEDER} & \textit{CVPR 2023} & 0.892 & 0.028 & 0.944 & 0.850 & 0.810 & 0.032 & 0.892 & 0.715 & 0.802 & 0.070 & 0.870 & 0.775 & 0.842 & 0.046 & 0.900 & 0.808 \\
    FGANet\cite{Zhai2023FGANet} & \textit{NeurIPS 2023} & 0.891 & 0.030 & 0.945 & 0.838 & 0.803 & 0.032 & 0.894 & 0.708 & 0.800 & 0.070 & 0.865 & 0.769 & 0.837 & 0.047 & 0.891 & 0.800 \\
    FocusDiff\cite{Zhao2024FocusDiffuser} & \textit{ECCV 2024} & 0.890 & 0.028 & 0.938 & 0.843 & 0.820 & 0.031 & 0.897 & 0.730 & 0.812 & 0.069 & 0.883 & 0.772 & 0.850 & 0.044 & 0.902 & 0.810 \\
    FSEL\cite{Sun2024FSEL} & \textit{ECCV 2024} & 0.893 & 0.029 & 0.941 & 0.847 & 0.822 & 0.032 & 0.891 & 0.722 & 0.816 & 0.069 & 0.881 & 0.779 & 0.847 & 0.045 & 0.901 & 0.807 \\
    \midrule
    \multicolumn{18}{c}{\cellcolor{subheadercolor}\textbf{\textit{Transformer-based Methods}}} \\
    \midrule
    HitNet\cite{Hu2023HitNet} & \textit{AAAI 2023} & \secondbest{0.907} & 0.024 & 0.944 & 0.861 & 0.847 & 0.027 & 0.922 & 0.790 & 0.834 & 0.060 & 0.892 & 0.791 & 0.858 & 0.042 & 0.911 & 0.825 \\
    VSCode\cite{luo2024vscode} & \textit{CVPR 2024} & - & - & - & - & \secondbest{0.869} & 0.023 & 0.931 & 0.806 & \secondbest{0.873} & 0.046 & \secondbest{0.925} & 0.844 & \secondbest{0.891} & 0.032 & \secondbest{0.934} & \best{0.863} \\
    CamoFocus \cite{Khan2024CamoFocus} & \textit{WACV 2024} & 0.906 & \secondbest{0.023} & \secondbest{0.953} & \secondbest{0.869} & 0.868 & \secondbest{0.022} & \secondbest{0.931} & \secondbest{0.818} & 0.870 & \secondbest{0.044} & 0.924 & \best{0.861} & 0.886 & \secondbest{0.031} & 0.932 & \secondbest{0.862} \\
    \midrule
    \rowcolor{ourscolor}
    \textit{Samba+} & \textit{Ours} & \best{0.920} & \best{0.018} & \best{0.954} & \best{0.879} & \best{0.886} & \best{0.019} & \best{0.937} & \best{0.822} & \best{0.882} & \best{0.042} & \best{0.927} & \secondbest{0.853} & \best{0.898} & \best{0.029} & \best{0.935} & 0.858 \\
    \bottomrule
  \end{tabular}
  \label{tabel_cod}
  \vspace{-4mm}
\end{table*}

\begin{figure}[t!]
    \centering
    \captionsetup{skip=5pt}
    \includegraphics[width=1\linewidth]{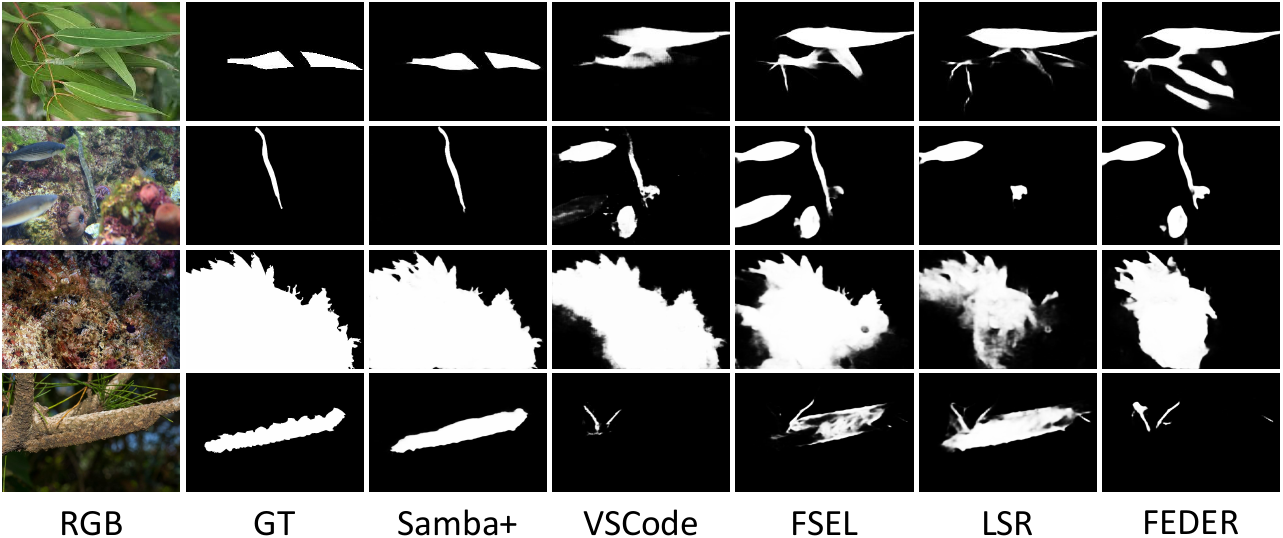}
    \caption{Visual comparison against SOTA COD methods.}
    \label{fig_cod}
    \vspace{-5mm}
\end{figure}

\begin{table}[t]

  \definecolor{ourscolor}{HTML}{FEF7F2} 
  
  \centering
  \small %
  \caption{Results on skin lesion segmentation (SLS). The best and second-best results are highlighted in \best{red} and \secondbest{blue}.}
  \vspace{-2mm}
  \setlength\tabcolsep{1.6pt} 
  \renewcommand{\arraystretch}{0.8} 
  \begin{tabular}{l| cccc| cccc}
    \toprule
    \multirow{2.5}{*}{Method} & \multicolumn{4}{c}{ISIC17\cite{Berseth2017ISIC}} & \multicolumn{4}{|c}{ISIC18\cite{Codella2019ISIC} } \\
    \cmidrule(lr){2-5} \cmidrule(lr){6-9}
    & \textit{IoU}$\uparrow$ & \textit{DSC}$\uparrow$ & \textit{Acc}$\uparrow$ & \textit{Spe}$\uparrow$ & \textit{IoU}$\uparrow$ & \textit{DSC}$\uparrow$ & \textit{Acc}$\uparrow$ & \textit{Spe}$\uparrow$ \\
    \midrule
    \text{TransFuse}$_{21}$   & 79.21 & 88.40 & 96.17 & 97.98 & 80.63 & 89.27 & 94.66 & 95.74 \\
    \text{EGE-UNet}$_{23}$    & 79.81 & 88.77 & 96.19 & 97.62 & 80.94 & 89.46 & 94.92 & 97.01 \\
    \text{VM-UNet}$_{24}$     & 80.23 & 89.03 & 96.29 & 97.58 & 81.35 & 89.71 & 94.91 & 96.13 \\
    \text{VM-UNetV2}$_{24}$   & \secondbest{82.34} & \secondbest{90.31} & 96.70 & 97.67 & \secondbest{81.37} & \secondbest{89.73} & \secondbest{95.06} & \secondbest{97.13} \\
    \text{UNetV2}$_{25}$     & 82.18 & 90.22 & \secondbest{96.78} & \secondbest{97.94} & 80.71 & 89.32 & 94.86 & 96.94 \\
    \midrule
    \rowcolor{ourscolor}
    \textit{Samba+} & \best{82.90} & \best{90.65} & \best{96.86} & \best{98.05} & \best{81.91} & \best{90.05} & \best{95.25} & \best{97.44} \\
    \bottomrule
  \end{tabular}
  \label{tabel_sls}
  \vspace{-5mm}
\end{table}

\subsection{Further Applications}
We validate \textit{Samba+}, the enhanced version of our framework, on camouflaged object detection and skin lesion segmentation to demonstrate its ability to model continuous spatial structures. 

\noindent
\textbf{Camouflaged Object Detection (COD).}
The evaluation is conducted on four benchmark datasets, \colorbox{gray!20}{CAMO}~\cite{Le2019Anabranch}, \colorbox{gray!20}{CHAMELEON}~\cite{Skurowski2018Chameleon}, \colorbox{gray!20}{COD10K}~\cite{Fan2020SINet}, and \colorbox{gray!20}{NC4K}~\cite{Lv2021LSR}, following standard training and testing protocols~\cite{Ji2023Deep, Mei2021Distraction}. We adopt four common metrics: mean absolute error (\textit{M}), adaptive F-measure (\(F_\beta\))~\cite{Margolin2014How}, mean E-measure (\(E_\phi\))~\cite{Fan2021Cognitive}, and structure measure ($S_m$)~\cite{fan2017structure}. As summarized in Table~\ref{tabel_cod}, \textit{Samba+} consistently surpasses recent methods across all datasets. Qualitative comparisons in Fig.~\ref{fig_cod} further show more complete object regions and sharper contours, especially under challenges such as occlusion, background clutter, and ambiguous boundaries.

\begin{figure}[t!]
    \centering
    \captionsetup{skip=5pt}
    \includegraphics[width=1\linewidth]{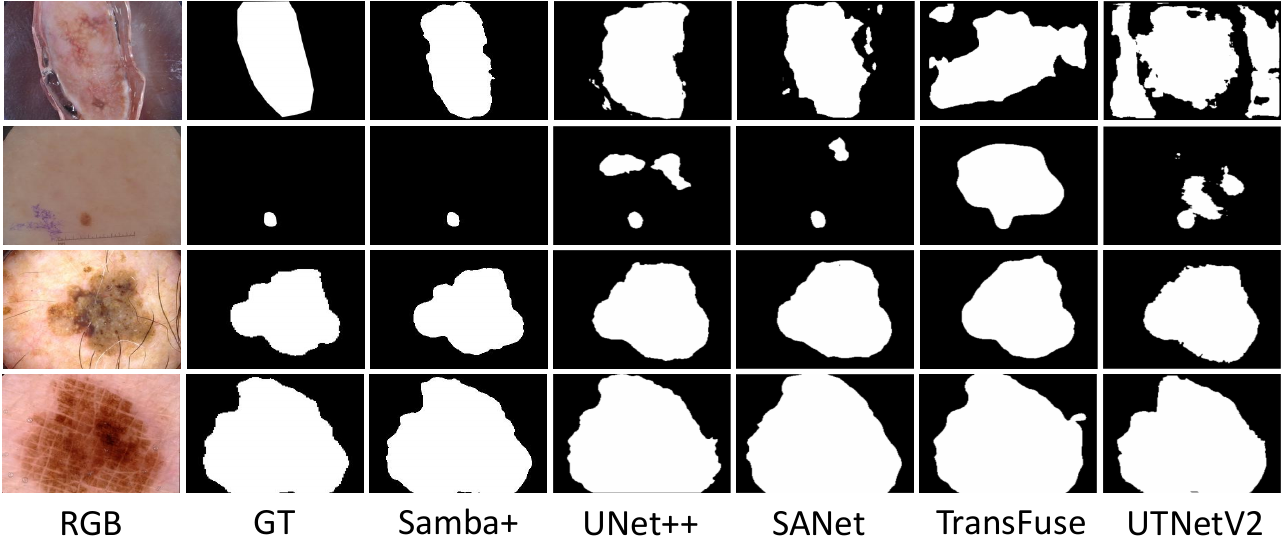}
    \caption{Visual comparison against SOTA SLS methods.}
    \label{fig_sls}
    \vspace{-3mm}
\end{figure}

\noindent
\textbf{Skin Lesion Segmentation (SLS).}
The evaluation is conducted on two benchmark datasets, \colorbox{gray!20}{ISIC2017}~\cite{Berseth2017ISIC} and \colorbox{gray!20}{ISIC2018}~\cite{Codella2019ISIC}, following standard training and testing protocols~\cite{Wei2021Shallow}. We compare our framework with five existing SOTA methods~\cite{Zhang2021TransFuse, Peng2025UNetV2,Ruan2023EGEUNET,Ruan2024VMUNET,Zhang2024VMUNETV2}. Evaluation metrics include mean IoU (IoU), Dice coefficient (DSC), accuracy (Acc), and specificity (Spe), where higher values indicate better performance. As shown in Table~\ref{tabel_sls}, \textit{Samba+} consistently achieves the best performance on both datasets. Visual results in Fig.~\ref{fig_sls} further show more precise boundaries and stronger spatial integrity.


\section{Conclusion}
\label{sec:Conclusion}
In this article, we are the first to adapt state space models to SOD tasks, proposing a new unified framework based on the pure Mamba architecture, named saliency Mamba (\ourmodel), which can flexibly handle general SOD tasks.
To overcome the ``task-specific'' deficiency in previous SOD research, we develop \textit{Samba+}, a more unified model by jointly training \textit{Samba} across heterogeneous SOD tasks. 
Extensive experiments demonstrate that \textit{Samba} independently achieves SOTA performance on six SOD tasks across 22 datasets, and \textit{Samba+} surpasses \textit{Samba} on all tasks and datasets within a single trained model, meanwhile maintaining lower computational cost. Notably, \textit{Samba} is the first systematic adaptation of state space models to SOD, demonstrating the effectiveness of introducing Mamba into the SOD domain. Building upon this, \textit{Samba+} is jointly trained on nearly all existing SOD datasets, pushing the framework closer to a more general and unified solution for SOD. With its modality-shared Siamese encoder and modality-adaptive converter design, \textit{Samba+} captures modality-invariant saliency cues more effectively, granting it strong generalization capability. With further validation on COD and SLS, it indicates that \textit{Samba+} could serve as a versatile framework in the near future for saliency-related downstream tasks that require continuous spatial reasoning.


\bibliographystyle{IEEEtran}
\bibliography{main} %
\vspace{-10mm}

\vfill
\end{document}